\documentclass[journal, onecolumn, 11pt, draftcls]{IEEEtran}
\usepackage{cite}
\usepackage{amsmath,amssymb,amsthm}
\usepackage{graphicx}
\usepackage{xcolor}
\usepackage{url}
\usepackage{subfigure}
\usepackage{balance}
\usepackage{caption}

\newcommand{\mbf}[1]{\mathbf{#1}}
\newcommand{\mcal}[1]{\mathcal{#1}}
\newcommand{\la}{\lambda}
\newcommand{\bs}{\backslash}

\newcommand{\vertiii}[1]{{\left\vert\kern-0.25ex\left\vert\kern-0.25ex\left\vert #1 
    \right\vert\kern-0.25ex\right\vert\kern-0.25ex\right\vert}}

\newcommand{\Etilde}{\widetilde{E}}

\newcommand{\Sighat}{\mbf{\widehat{\Sigma}}}
\newcommand{\Sigstar}{\mbf{\Sigma}^{*}}

\newcommand{\JRelax}{\widehat{\mbf{J}}^{\text{Relax}}}

\newcommand{\KRelax}{\widehat{\mbf{K}}^{\text{Relax}}}

\newcommand{\Khat}{\widehat{\mbf{K}}}
\newcommand{\Kstar}{\mbf{K}^{*}}
\newcommand{\Jstar}{\mbf{J}^{*}}
\newcommand{\Deltahat}{\widehat{{\Delta}}}
\newcommand{\Del}{{\Delta}}

\newcommand{\be}{\begin{equation}}
\newcommand{\ee}{\end{equation}}
\newcommand{\ALGN}[1]{\begin{align*}#1\end{align*}}
\newcommand{\ALGNN}[1]{\begin{align}#1\end{align}}

\newtheorem{theorem}{Theorem}
\newtheorem{lemma}{Lemma}

\DeclareMathOperator*{\argmin}{arg\,min}

\DeclareMathOperator*{\trace}{trace}

\usepackage{authblk}

\title{Marginal Likelihoods for Distributed Parameter Estimation of Gaussian Graphical Models}

\author{Zhaoshi~Meng, ~\IEEEmembership{Student Member,~IEEE,} 
	Dennis~Wei, ~\IEEEmembership{Member,~IEEE,}
	Ami~Wiesel, ~\IEEEmembership{Member,~IEEE,} and
	Alfred~O.~Hero~III, ~\IEEEmembership{Fellow,~IEEE,}%
\thanks{This research was supported in part by ARO grant W911NF-11-1-0391 and ISF 786/11. }
\thanks{The material in this paper was presented in part at the Sixteenth International Conference on Artificial Intelligence and Statistics~\cite{meng2013distributed}, and the Fifth IEEE Workshop on Computational Advances in Multi-Sensor Adaptive Processing~\cite{meng2013marginal}.}
\thanks{Zhaoshi Meng and Alfred O. Hero III are with the Department of Electrical Engineering and Computer Science, University of Michigan, Ann Arbor, MI 48109 USA (email: mengzs,hero@umich.edu).}% %Tel: (734) 763-0564. FAX: 734-763-8041 }%
\thanks{Dennis Wei is with the IBM T.~J.~Watson Research Center, Yorktown Heights, NY 10598 USA (email: dwei@us.ibm.com).}
\thanks{Ami Wiesel is with the School of Computer Science and Engineering, The Hebrew University of Jerusalem, 91904 Jerusalem, Israel (email: ami.wiesel@huji.ac.il).}% Tel/Fax: 972-2-5494539}
}
\date{}

\begin{document}
\maketitle

\begin{abstract}
We consider distributed estimation of the inverse covariance matrix, also called the concentration or precision matrix, in Gaussian graphical models. Traditional centralized estimation often requires  global inference of the covariance matrix, which can be computationally intensive in large dimensions. 
Approximate inference based on message-passing algorithms,  on the other hand, can lead to unstable and biased estimation in loopy graphical models. 
 In this paper, we propose a general framework for distributed estimation based on a maximum marginal likelihood (MML) approach.  
 This approach computes local parameter estimates by maximizing marginal likelihoods defined with respect to data collected from local neighborhoods. 
 Due to the non-convexity of the MML problem, we 
introduce and solve
a convex relaxation.  The local estimates are then combined into a global estimate without the need for iterative message-passing between neighborhoods. 
The proposed algorithm is naturally parallelizable and computationally efficient, thereby making it suitable for high-dimensional problems. 
In the classical regime where the number of variables $p$ is fixed and the number of samples $T$ increases to infinity, the proposed estimator is shown to be asymptotically consistent and to improve monotonically as the local neighborhood size increases.  
In the high-dimensional scaling regime where both $p$ and $T$ increase to infinity, the convergence rate to the true parameters is derived and 
is seen to be comparable to centralized maximum likelihood estimation.  
Extensive numerical experiments demonstrate the improved performance of the two-hop version of the proposed estimator, which suffices to almost close the gap to the centralized maximum likelihood estimator at a reduced computational cost.

\end{abstract}


\section{Introduction}

Graphical models provide a principled framework for compactly characterizing dependencies among many random variables, represented as nodes in a network~\cite{lauritzen1996graphical, wainwright2008graphical}. Their sparse structure allows for efficient and distributed inference using message-passing algorithms such as loopy belief propagation (LBP), which makes them especially well-suited to large networks, such as sensor, social, and biological networks~\cite{liu12a, wiesel2012distributed, meng2012distributed}. Less well-studied, however, is the equally important task of distributed estimation of the parameters of a graphical model from data.  The goal of this work is to develop and analyze distributed methods for model parameter estimation.

In this paper we focus on Gaussian graphical models (GGM) with known graph structure, i.e, the pattern of edges is known.
Our approach can also be extended to more general graphical models, including discrete distributions. For GGMs, parameter estimation essentially reduces to (inverse) covariance estimation, and knowledge of the edge pattern imposes sparsity constraints on the inverse covariance matrix, also known as the concentration or precision matrix.
While the resulting GGM maximum likelihood (ML) parameter estimation problem is a convex optimization, solving it exactly for generally structured networks 
using centralized algorithms as in~\cite{banerjee2006convex, dahl2008covariance, friedman2008sparse} becomes impractical in large real-world networks 
where data collection and computational resources are limited.

%%%%%%%%%%%%%%%%%%%%%%%%%%%%%%%%%%%%%%%%%%%%%%%%%%%%%%
A natural approach toward distributed parameter estimation is to leverage methods for distributed marginal inference, such as LBP and its extensions.  The idea is to replace the objective function and its gradient in the ML estimation problem with approximations that can be computed through iterative message-passing.  
However, in many cases LBP may fail to converge or give good marginal estimates, and when it does converge, the resulting parameter estimates may be biased~\cite{malioutov2006walk, heinemann2012cannot}.

Another direction for distributed estimation is to consider a surrogate objective that decomposes into smaller problems that are locally parameterized. Then a distributed ML algorithm estimates the   
local parameters by processing local data with limited message passing. 
Some recent efforts along this direction~\cite{wiesel2012distributed,liu12a} have considered a pseudo-likelihood framework for exponential family distributions.

This paper proposes a general framework for distributed estimation based on \emph{marginal} likelihoods, as contrasted with pseudo-likelihoods.  Each node collects data within its extended neighborhood and independently forms a local estimate by maximizing a marginal likelihood.  To deal with the non-convexity of the maximum marginal likelihood (MML) estimation problem, we  %derive and consider solving 
formulate a convex relaxation of the problem. The resulting distributed estimator is computationally efficient, and involves minimal message passing. 
%%%%%%%%%%%%%%%%%%%%%%%%%%%%%%%%%

We analyze the mean squared error (MSE) of the proposed distributed estimator in both the classical asymptotic regime (fixed number of parameters $p$ and increasing number of samples $T \rightarrow \infty$), and also the high-dimensional regime where both $p$ and $T$ increase to infinity ($p, T \rightarrow \infty$). 
In the classical regime, the distributed estimator is shown to be asymptotically consistent.  Furthermore, the asymptotic error improves monotonically as the local neighborhood size increases.  In the high-dimensional regime,  we show that under certain conditions and proper scaling between $p$ and $T$, the proposed estimator achieves a comparable statistical convergence rate to the (more expensive) global ML estimator.

Our analytical results are supported by extensive numerical experiments on both synthetic and real-world data sets. In particular, we show that two-hop local information is sufficient for the proposed distributed estimator to match the performance of the more expensive centralized ML estimator. 
 The proposed estimator also improves significantly upon existing distributed estimators~\cite{liu12a, wiesel2012distributed}.  In terms of computation, the complexity of our estimator increases at most linearly with $p$ in most cases and can be further reduced through parallelization. In the case of a physical network implementation, the near-absence of message passing and long-distance communication is also an advantage.

We emphasize that the problem we consider is different from covariance selection~\cite{ravikumar2011high, rothman2008sparse, johnson2011high, friedman2008sparse}, in which the graph topology is not known \emph{a priori} and must be estimated in addition to the parameters.  
 To test our assumption of known graph structure, we also study the robustness of the proposed estimators against small model   (i.e. structure)  mismatch. Both theoretical analysis and numerical results show that the proposed distributed estimator is as robust as the centralized ML estimator.

The algorithm and some preliminary experimental results in the current paper were first presented in~\cite{meng2013distributed}. During the preparation of this extended version, a related and independent work~\cite{mizrahi2013efficient} has come to our attention. The authors of~\cite{mizrahi2013efficient} consider a distributed learning algorithm for general Markov random fields which works on local unions of cliques, generalizing nodes and edges in the Gaussian case. Asymptotic consistency is also discussed. 
After submission of this extended version, an anonymous reviewer made us aware of an earlier arXiv posting~\cite{massam2013distributed} by Massam and Wang that considered a version of our algorithm~\cite{meng2013distributed} for discrete graphical models. Focusing on the discrete case, they established novel asymptotic theory 
%(consistency and monotonicity of variance) 
that parallels the theory developed in the present paper for the Gaussian case.

The outline of the paper is as follows. 
In Section \ref{sec:bkgd}, we give a brief review of graphical models, centralized ML parameter estimation, and the difficulty of parameter estimation using traditional marginal inference techniques. In Section \ref{sec:estimatorGGM}, we propose a general approach to distributed estimation based on marginal likelihoods. In Section \ref{sec:analysis}, we provide extensive analysis of the convergence rates and robustness of the proposed estimator.   Section~\ref{sec:computation} discusses the computational complexity and implementation advantages of the estimator.   Numerical experiments are presented in Section \ref{sec:exp} and the paper concludes in Section \ref{sec:conclusion}. \\

\noindent\textbf{Notation.} Boldface upper case letters denote matrices and boldface lower case letters denote column vectors. Sets of single indices are denoted by calligraphic upper case letters.  The cardinality of a set $\mathcal{A}$ is denoted by $|\mathcal{A}|$ and the difference of two sets is denoted as $\mathcal A \backslash \mathcal B$. 
Following common notation, $\mbf{A}_{\mcal{M},\mathcal{N}}$ represents a submatrix of $\mbf{A}$ with rows indexed by $\mcal{M}$ and columns indexed by $\mcal{N}$.  We also make reference to irregular sets of index pairs such as the edge set $E$ of a graph, for which we use standard upper case letters. $\mbf{A}_{E}$ then refers to the vector of entries of $\mbf{A}$ indexed by $E$.  The standard inner product between two symmetric matrices is denoted as $\langle \mbf{A}, \mbf{B} \rangle$, i.e., 
%
%\[
$\langle \mbf{A}, \mbf{B} \rangle = \trace(\mbf{A} \mbf{B}) = \sum_{i,j} \mbf{A}_{i,j} \mbf{B}_{i,j}$.
%\]
We distinguish the following two norms for matrices: the induced $\ell_\infty / \ell_\infty$ norm 
%$\| \mbf{A} \|_{\infty/\infty} := \max_{i = 1,\ldots,p} \sum_{j=1}^p | \mbf{A}_{i,j} |$, 
$\vertiii{\mbf{A}}_{\infty} := \max_{i = 1,\ldots,p} \sum_{j=1}^p | \mbf{A}_{i,j} |$, 
and the element-wise $\ell_\infty$ norm $\| \mbf{A} \|_\infty := \max_{i,j = 1, \ldots,p} | \mbf{A}_{i,j} |$. $\la_{\max}(\mbf{A})$ and $\la_{\min}(\mbf{A})$ denote the maximum and minimum eigenvalues of matrix $\mbf{A}$, respectively. 

\section{Background}
\label{sec:bkgd}

We begin by providing background on graphical models and their statistical inference. We refer the reader to~\cite{lauritzen1996graphical, wainwright2008graphical} for a detailed treatment. 

\subsection{Gaussian Graphical Models}

Consider a $p$-dimensional random vector $\mbf{x}$ following a graphical model with respect to an undirected graph $\mcal{G} = (V,E)$, where $V = \{ 1, \ldots, p \}$ is a set of nodes corresponding to elements of $\mbf{x}$ and $E$ is a set of edges connecting nodes.
The vector $\mbf{x}$ satisfies the Markov property with respect to $\mcal{G}$ if for any pair of nonadjacent nodes in $\mcal{G}$, the corresponding pair of variables in $\mbf{x}$ are conditionally independent given the remaining variables. 

If the vector $\mbf{x}$ follows a multivariate Gaussian distribution, the corresponding model is called a Gaussian graphical model (GGM).  We assume without loss of generality that $\mbf{x}$ has zero mean.  Then the probability density function can be written in canonical form in terms of the concentration matrix $\mbf{J}$ as follows: 
\ALGNN{
p(\mbf{x} ; \mbf{J}) = (2 \pi)^{-p/2} (\det \mbf{J})^{1/2} \exp \left( -\frac{1}{2} \mbf{x}^{T} \mbf{J} \mbf{x} \right).
\label{eq:GGM}
}  
The Markov property manifests itself in a simple way through the sparsity pattern of $\mbf{J}$:
\ALGNN{
\mbf{J}_{i,j} = 0 \text{ for all }   i \neq j, (i,j) \notin E. \label{eq:markov}
}

\subsection{Maximum Likelihood Parameter Estimation for GGMs}

Estimating the parameters of a graphical model from sample data is the first step for many applications. For Gaussian graphical models this reduces to estimating the non-zero elements of the concentration matrix $\mbf{J}$
(including the diagonal elements).  Defining 
\ALGNN{
\Etilde := E \cup \{ (i,i)\}_{i = 1}^{p}
\label{eq:Etilde}
}
as the index set for these non-zero elements,  
the centralized global maximum likelihood (GML) estimation problem can be formulated as~\cite{lauritzen1996graphical}:
\ALGNN{
\begin{split}
\widehat{\mbf{J}}^{\mathrm{GML}} & = \argmin_{\mbf{J}}  \ \langle \Sighat, \mbf{J}  \rangle - \log \det \mbf{J} \\
\text{ s.t. } & \ \mbf{J}_{j,k} = 0 \quad \forall \; (j,k) \notin \Etilde \\
&\ \mbf{J} \succeq \mbf{0},
\end{split}
\label{eq:gml}
}
where 
\[
\Sighat = \frac{1}{T} \sum_{t=1}^{T} \mbf{x}(t) \mbf{x}(t)^{T}
\]
is the sample covariance matrix and $\mbf{x}(1),\ldots,\mbf{x}(T)$ are i.i.d.~samples of $\mbf{x}$. 

The GML problem \eqref{eq:gml} is a convex log-determinant-regularized semidefinite program ($\log \det$-SDP) with respect to $\mbf{J}_{\Etilde}$ and various gradient-based algorithms can be applied to solve this problem, many of which have specialized implementations on graphs, \emph{e.g.}~iterative proportional fitting (IPF) \cite{wainwright2008graphical}, chordally-embedded Newton's method \cite{dahl2008covariance}, etc. 
The standard gradient descent algorithm for solving problem~\eqref{eq:gml} has the following update rule at each iteration:
\ALGNN{
\begin{split}
\widehat{\mbf{J}}^{(t+1)}_{i,j} & \leftarrow \widehat{\mbf{J}}^{(t)}_{i,j} - \gamma \cdot \nabla {\ell}(\widehat{\mbf{J}}^{(t)})_{i,j} \\
& = 
\begin{cases}
\widehat{\mbf{J}}^{(t)}_{i,j} - \gamma \cdot \left( 2 \Sighat_{i,j} - 2 (\widehat{\mbf{J}}^{(t)})^{-1}_{i,j} \right),  \quad i \neq j \\
\widehat{\mbf{J}}^{(t)}_{i,j} - \gamma \cdot \left(\Sighat_{i,j} -  (\widehat{\mbf{J}}^{(t)})^{-1}_{i,j} \right),  \quad  i = j
\end{cases}
\end{split}
\label{eq:graddesc}
}
where $\ell(\mbf{J})$ is the GML objective function and $\nabla \ell(\mbf{J})$ denotes its gradient, $\gamma$ is the step-size, and we have used the facts $\frac{\partial \log \det \mbf{X}}{\partial \mbf{X}_{i,j}} = 2 (\mbf{X}^{-1})_{i,j}$ for $i \neq j$ and $\frac{\partial \log \det \mbf{X}}{\partial \mbf{X}_{i,i}} = (\mbf{X}^{-1})_{i,i}$ for symmetric matrices~\cite{Petersen06thematrix}.
The obvious difficulty is the global matrix inversion involved in computing the gradient at each step, 
whose computational cost is cubic in the number of variables for generally structured models.

Given the expense of the matrix inversion in~\eqref{eq:graddesc}, an alternative is to consider  distributed message-passing algorithms, such as loopy belief propagation (LBP), an iterative message-passing algorithm for inference of marginal distributions. When applied to tree-structured graphs, LBP yields exact marginals. Unfortunately, this does not hold for loopy graphs in general~\cite{murphy1999loopy}. 
For Gaussian models, 
many \emph{sufficient} conditions exist for Gaussian LBP to converge, such as diagonal dominance, walk-summablility, pairwise normalizability, etc.~\cite{malioutov2006walk}. However, when these sufficient conditions do not hold, 
Gaussian LBP can be divergent, or it may converge to degenerate, unnormalized marginal distributions.
A recent work~\cite{pacheco2012minimization} uses the method of multipliers to improve the convergence behavior of Gaussian LBP for some less ill-conditioned models.
However, 
even if LBP converges, 
its final estimate is not guaranteed to be consistent. 
For discrete graphical models, the authors of \cite{heinemann2012cannot} show that many models are in principle not learnable through LBP, 
which implies that an estimator based on LBP inference is inevitably biased for a subset of models. Similar drawbacks also hold when using  other approximate inference techniques, for example, tree-reweighted BP~\cite{wainwright2006estimating}. 
The above difficulties of parameter estimation using traditional marginal inference techniques motivate us to consider a different distributed framework for parameter estimation, as introduced in the next section.

%%%%%%%%%%%%%%%%%%%%%%%%%%%%%%%%%%%%%

\section{Distributed Estimation in GGMs}
\label{sec:estimatorGGM}

Our  framework avoids the weakness of LBP and other message passing approaches to distributed estimation of GGMs. The proposed distributed algorithm collects all the data samples from within each neighborhood and computes a local parameter estimate. A global estimate of the parameter (e.g. precision matrix $\mbf{J}$) is then formed by combining these local estimates with a simple, single pass aggregation rule.

\begin{figure*}[t]
	\begin{subfigure}[2D lattice and two-hop neighborhood $\mcal{N}_{i}$]{
	\label{subfig:grid}
		\centering
		\includegraphics[width=0.30\textwidth]{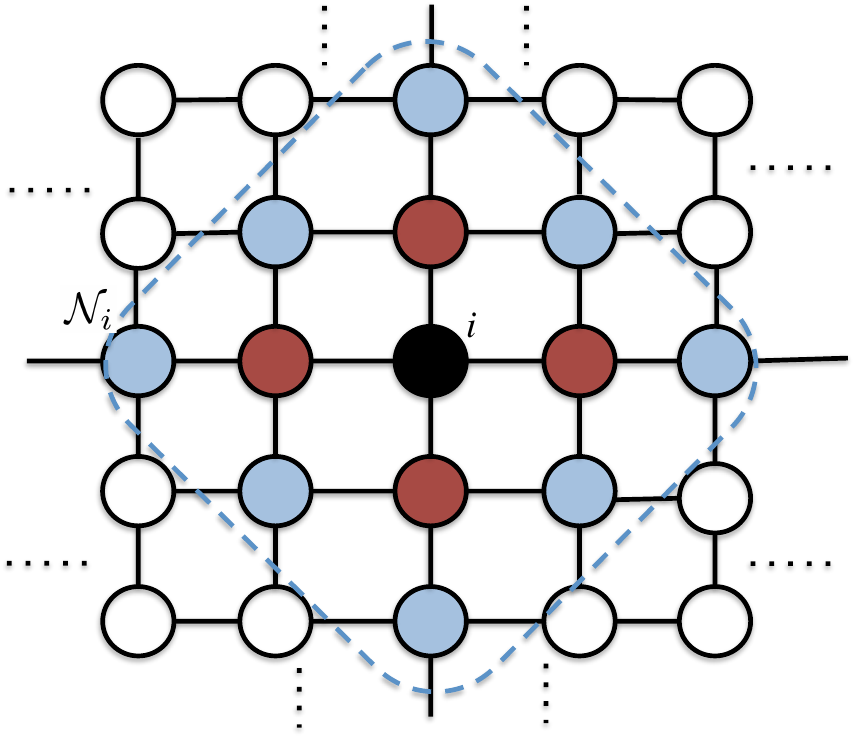}
	}
	\end{subfigure}
%	~ 
	\begin{subfigure}[A general graph and two-hop neighborhood $\mcal{N}_{i}$]{
	\label{subfig:general}
		\centering
		\includegraphics[width=0.30\textwidth]{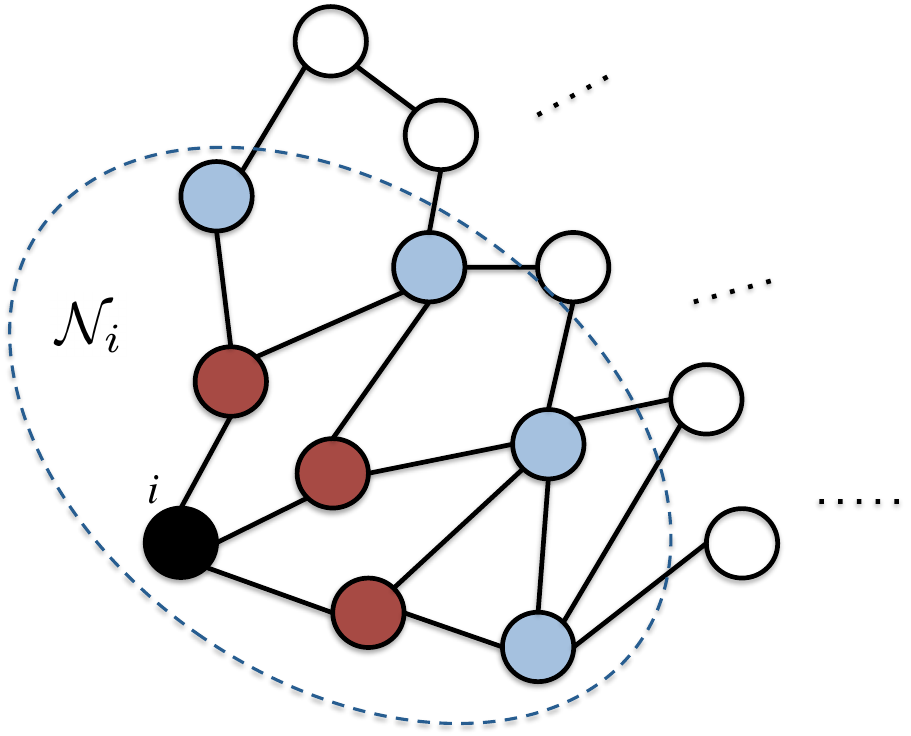}
	}
	\end{subfigure}
%	~ 
	{\subfigcapskip = -10pt
	\begin{subfigure}[Local relaxations (one-hop (left) and two-hop (right)). Dotted lines denote fill-in edges.]{
	\label{subfig:localproblems}
		\centering
		\raisebox{0.3in}{\includegraphics[width=0.30\textwidth]{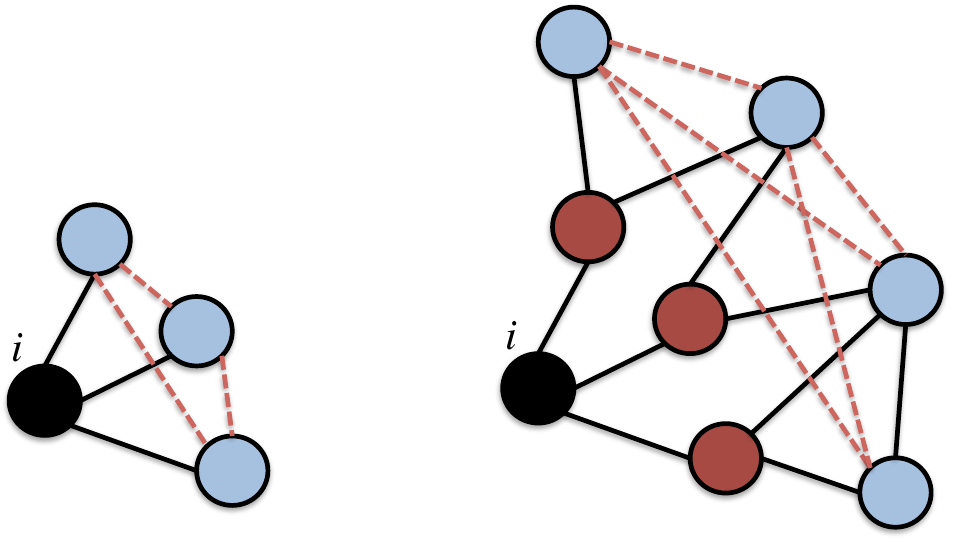}}
	}
	\end{subfigure}
	}
\caption{Illustration of defined sets in the proposed local relaxation of MML. In (a) and (b) we show two different graphs, in which the two-hop neighborhood $\mcal{N}$ for node $i$ is indicated with dashed contours. The buffer set variables $\mbf{x}_{\mcal{B}}$ and the protected set variables $\mbf{x}_{\mcal{P}}$ (excluding node $i$ itself) are colored blue and red, respectively. For the graph in (b), we illustrate the one-hop and two-hop local relaxations in (c). The dashed red lines in (c) denote the fill-in edges due to relaxation. 
}
\label{fig:illustrations}
\end{figure*}
%%%%%%%%%%%%%%%%%%%%%%%%%%%%%%%%%%%%%%%%

\subsection{Marginal Likelihood Maximization}
\label{sec:MMLE}

We consider estimating local parameters by maximizing \emph{marginal likelihood} functions in neighborhoods around each node. Define the index set for \emph{immediate neighbors} of node $i$ as 
\ALGNN{
\mcal{I}_{i} := \{ j \ | \ (i,j) \in E\},
}
and consider a neighborhood indexed by a set $\mcal{N}_{i}$ containing at least the node $i$ itself and its immediate neighbors $\mcal{I}_{i}$. Let $\mbf{K}$ denote the concentration matrix corresponding to the marginal distribution over the variables $\{ \mbf{x}_{j}, j \in {\mcal{N}_{i}} \}$ in the neighborhood, and let 
%
%\[
%\mbf{S}^{i} := 
$\Sighat_{\mcal{N}_{i}, \mcal{N}_{i}} = \frac{1}{T} \sum_{t=1}^{T} \mbf{x}_{\mcal{N}_{i}}(t) \mbf{x}_{\mcal{N}_{i}}(t)^{T}$
%\]
%
be the marginal sample covariance matrix.
The maximum marginal likelihood (MML) estimation problem in neighborhood $\mcal{N}_{i}$ can be formulated as:
% a joint optimization of $\mbf{K}$ and $\mbf{J}$:
\ALGNN{
\begin{split}
\widehat{\mbf{K}}^{i,\text{MML}} & = \argmin_{\mbf{K}, \mbf{J} }  \ \langle \Sighat_{\mcal{N}_{i}, \mcal{N}_{i}}, \mbf{K}  \rangle - \log \det \mbf{K} \\
\text{ s.t. } & \ \mbf{K} = \left[ \left( \mbf{J}^{-1} \right)_{\mcal{N}_{i}, \mcal{N}_{i}} \right]^{-1},  \\
&\ \mbf{J}_{j,k} = 0 \quad \forall \; (j,k) \notin \Etilde, \\
&\ \mbf{J} \succeq \mbf{0}, 
\end{split}
 \label{eq:localml}
}
where the first constraint represents the marginalization relationship between $\mbf{K}$ and the global precision matrix $\mbf{J}$, and the second line of constraints reflects the global sparsity constraints.   We index the nodes in the MML problem~\eqref{eq:localml} in the same way as in the GML problem~\eqref{eq:gml}. (For example, if $\mcal{N}_1 = \{1, 3, 6\}$, the rows and columns of $\mbf{K}$ are indexed by $\{1, 3, 6\}$ and not re-indexed to $\{1, 2, 3\}$.)

The difficulty with direct application of MML is that problem \eqref{eq:localml} is in general a non-convex optimization with respect to $\mbf{K}$ and $\mbf{J}$. 
The non-convexity arises from the coupling of the nonlinear marginalization constraint linking $\mbf{K}$ to $\mbf{J}$ and the sparsity constraints on $\mbf{J}$.
As a surrogate, we derive next a convex relaxation of the MML estimation problem. %as a surrogate.  

\subsection{Convex Relaxation of MML} 
\label{sec:RMMLE}

We apply the Schur complement identity to the marginalization constraint in \eqref{eq:localml}, yielding 
\ALGNN{
\mbf{K} = \mbf{J}_{\mcal{N}, \mcal{N}} - \mbf{J}_{\mcal{N}, \mcal{N}^{C}} \cdot \left[ \mbf{J}_{\mcal{N}^{C}, \mcal{N}^{C}} \right]^{-1} \cdot \mbf{J}_{\mcal{N}^{C}, \mcal{N}},
\label{eq:schur}
}
where $\mcal{N}^{C}$ is the complementary set to $\mcal{N}$,  and we have dropped the subscript $i$ to simplify notation.
Define the \emph{buffer set} $\mcal{B} \subset \mcal{N}$ as the set of all variables in $\mcal{N}$ that have immediate neighbors in the complement $\mcal{N}^{C}$,
\ALGNN{
\mcal{B} := \{ j \ | \ j \in \mcal{N} \text{ and } \mcal{I}_{j} \cap \mcal{N}^{C} \neq \emptyset \}.
\label{eq:setB}
}
The difference set between $\mcal{N}$ and $\mcal{B}$ is referred to as the \emph{protected set} $\mcal{P} := \mcal{N} \bs \mcal{B}$. 
The buffer and protected sets are illustrated in Figure~\ref{subfig:grid} and~\ref{subfig:general}. Due to the Markov property, we have
$\mbf{J}_{\mcal{P},\mcal{N}^{C}} = \mbf{0}.$
Decomposing $\mcal{N}$ into $\mcal{B}$ and $\mcal{P}$ then reveals the sparsity pattern of $\mbf{K}$ using \eqref{eq:schur}:
\ALGNN{
\mbf{K} & = \mbf{J}_{\mcal{N}, \mcal{N}} - 
\begin{bmatrix}
	\begin{array}{c}
	\mbf{0} \\
	\mbf{J}_{\mcal{B}, \mcal{N}^{C}} 
	\end{array}
\end{bmatrix}
\left[ \mbf{J}_{\mcal{N}^{C}, \mcal{N}^{C}} \right]^{-1} 
\begin{bmatrix}
	\begin{array}{cc}
	\mbf{0}, \  \mbf{J}_{\mcal{N}^{C}, \mcal{B}}
	\end{array}
\end{bmatrix}, \notag \\
& = \mbf{J}_{\mcal{N}, \mcal{N}} - 
\begin{bmatrix}
	\begin{array}{cc}
	\mbf{0} & \mbf{0} \\
	\mbf{0} & \mbf{J}_{\mcal{B}, \mcal{N}^{C}}\left[ \mbf{J}_{\mcal{N}^{C}, \mcal{N}^{C}} \right]^{-1} \mbf{J}_{\mcal{N}^{C}, \mcal{B}}
	\end{array}
\end{bmatrix} 	\notag
}
and hence 
\ALGNN{
\mbf{K}_{\mcal{P},\mcal{P}} & = \mbf{J}_{\mcal{P},\mcal{P}}, \ \mbf{K}_{\mcal{P},\mcal{B}}  = \mbf{J}_{\mcal{P},\mcal{B}},  \label{eq:KequalsJ} \\
\mbf{K}_{\mcal{B},\mcal{B}} & =  \mbf{J}_{\mcal{B},\mcal{B}} - \mbf{J}_{\mcal{B}, \mcal{N}^{C}}\left[ \mbf{J}_{\mcal{N}^{C}, \mcal{N}^{C}} \right]^{-1} \mbf{J}_{\mcal{N}^{C}, \mcal{B}}.	\label{eq:KequalsSchur}
} 

An important observation from  \eqref{eq:KequalsJ} is that in the rows and columns indexed by the protected set $\mcal{P}$, the sparsity pattern of $\mbf{J}_{\mcal{N},\mcal{N}}$ is entirely preserved and the local parameters are equal to the global ones. On the other hand, the sparsity pattern in the ``buffer submatrix'' $\mbf{K}_{\mcal{B},\mcal{B}}$ is in general modified from $\mbf{J}_{\mcal{B}, \mcal{B}}$ due to the fill-in term, \emph{i.e.}, the second term in \eqref{eq:KequalsSchur}. 

Based on these observations, we now specify a relaxed set of constraints on the marginal concentration matrix $\mbf{K}$. First denote the set of all local edges that are not affected by the fill-in term in \eqref{eq:KequalsSchur} as 
\ALGNN{
E^{\mathrm{Prot}} := \Etilde \cap \left\{ 
\{ \mcal{P} \times \mcal{P} \} \cup \{ \mcal{P} \times \mcal{B} \} \cup \{ \mcal{B} \times \mcal{P} \} \right\},
}
where the superscript stands for ``protected''.  We then add to $E^{\mathrm{Prot}}$ all index pairs $\mcal{B} \times \mcal{B}$ that could potentially be affected by fill-in in \eqref{eq:KequalsSchur}, resulting in a \emph{relaxed edge set} $R$ (see Figure~\ref{subfig:localproblems} for illustrations): 
\ALGNN{
R & = 
E^{\mathrm{Prot}} \cup \{ \mcal{B} \times \mcal{B} \}. 
\label{eq:setS}
}
In light of \eqref{eq:KequalsJ} and \eqref{eq:KequalsSchur}, any feasible marginal concentration matrix $\mbf{K}$ for the MML estimation problem \eqref{eq:localml} is guaranteed to be supported only on the set $R$. Therefore we can relax the feasible set and formulate the following relaxation of~\eqref{eq:localml} at each node $i$, called the relaxed MML (RMML) problem: 
\ALGNN{
\begin{split}
\widehat{\mbf{K}}^{i,\text{Relax}} & = \argmin_{\mbf{K}}  \ \langle \Sighat_{\mcal{N}_{i}, \mcal{N}_{i}}, \mbf{K}  \rangle - \log \det \mbf{K} \\
\text{ s.t. } & \ \mbf{K}_{j,k} = 0 \quad \forall \; (j,k) \notin R \\
&\ \mbf{K} \succeq \mbf{0}.
\end{split}
\label{eq:localrelax}
}
The above RMML problem is a convex optimization with respect to $\mbf{K}$ and has the same form as the global MLE problem \eqref{eq:gml} but over matrices of much lower dimension. 

After solving the RMML estimation problems as surrogates to estimate local parameters, a global estimate of the concentration matrix can then be constructed by extracting a subset of parameters from each local estimate and concatenating them. Specifically, we  extract the local parameter estimates indexed by 
\ALGNN{
L_{i} := \{ (j,k) \in \Etilde \ | \ j = i \},
\label{eq:setL}
}
\emph{i.e.}, the non-zero entries in the $i$th row of $\mbf{J}$. We refer to the parameters indexed by $L_{i}$ as the \emph{row parameters} for node $i$. From~\eqref{eq:KequalsJ}, when there are no sampling errors, i.e. $T \rightarrow \infty$, the marginal and global concentration matrices are guaranteed  to share the same parameters in $L_{i}$.
Therefore our global estimate of $\mbf{J}$ is formed by concatenating local solutions of~\eqref{eq:localrelax}:  
\ALGNN{
\widehat{\mbf{J}}_{L_{i}}^{\mathrm{Relax}} \leftarrow \widehat{\mbf{K}}_{L_{i}}^{i,\mathrm{Relax}}, \ \text{ for } i = 1,\ldots, p.
\label{eq:estimatorRMMLE}
}

The proposed RMML framework is very general and applies to many possible choices of local neighborhoods, which include, e.g., nearest neighbors, second-order nearest neighbors, or, in general, $k$-th order nearest neighbors of a node $i$.
In the following subsections, we consider one- and two-hop neighborhoods. The absence of sampling errors is still assumed, i.e. $T \rightarrow \infty$.

\subsection{Case I: One-hop Estimator}
\label{sec:onehop}

We first consider a first-order (i.e., one-hop) neighborhood consisting of node $i$ and its immediate neighbors $\mcal{I}_{i}$, i.e., $\mcal{N}_{i} = \{ i \} \cup\mcal{I}_{i}$. Generically in the worst case where the immediate neighbors are all buffer nodes, we have $\mcal{B}_{i} = \mcal{I}_{i}$, and $\mcal{P}_{i} = \{i\}$. The fill-in term in \eqref{eq:KequalsSchur} affects the submatrix $\mbf{K}_{\mcal{I}_{i},\mcal{I}_{i}}$, leaving only the first row and column untouched. In this case, since $i$ is by definition connected to all elements in $\mcal{I}_{i}$, the relaxed edge set $R_{i}$ defined in \eqref{eq:setS} includes all possible pairs (see leftmost graph of Figure~\ref{subfig:localproblems} for an illustration): 
%\ALGN{
$R_{i}^{\mathrm{1hop}} =  \mcal{N}_{i} \times \mcal{N}_{i}$.
%}

The solution to the relaxed MML problem \eqref{eq:localrelax} 
for the first-order neighborhood is simply the inverse of the local sample covariance,% (assuming enough samples for invertibility),
\ALGNN{
\widehat{\mbf{K}}^{i,\mathrm{1hop}}  
& = \left( \Sighat_{\mcal{N}_{i}, \mcal{N}_{i}} \right)^{-1}.	\label{eq:onehoplocal}
}
The global estimate is obtained by combining the local one-hop estimates as in \eqref{eq:estimatorRMMLE}. 

In the one-hop case, the proposed relaxed MML estimator reduces to the \emph{LOC} estimator in~\cite{wiesel2012distributed}. 
As shown in~\cite{wiesel2012distributed}, this estimator is also equivalent to the pseudolikelihood estimator~\cite{liang2008asymptotic} without symmetry constraints, and the covariance selection procedure in~\cite{friedman2008sparse} when the graph is known.

\subsection{Case II: Two-hop Estimator}
\label{sec:twohop}

We next consider a second-order neighborhood (two-hop), $\mcal{N}_{i}$ that includes nodes that are reachable from node $i$ within two hops. In this setting, the worst-case protected set is given by $\mcal{P}_{i} = \{ i \} \cup \mcal{I}_{i}$ and the buffer set $\mcal{B}_{i} = \mcal{N}_{i} \bs \mcal{P}_{i}$ consists of all nodes that are exactly two hops away from the $i$th node. 
Hence $\mcal{B}_{i}$ can be thought of as the set of \emph{second-hop} nodes. 
In the two-hop case, the protected edge set $E^{\mathrm{Prot}}$ includes not only edges between node $i$ and its immediate first-hop neighbors, but also edges between first-hop neighbors and between first- and second-hop neighbors (see Figure~\ref{subfig:localproblems} for an illustration).

Unlike in the one-hop case, the two-hop problem~\eqref{eq:localrelax} does not admit a general closed-form solution. However, as mentioned before, Eq.~\eqref{eq:localrelax} can be solved using efficient algorithms for semidefinite programming. A global estimate is obtained as before by combining row parameter estimates \eqref{eq:estimatorRMMLE}.

\subsection{Symmetrization of RMML Estimator}
\label{sec:localaverage}

When $\Sighat$ is estimated from finite sample sizes, the local estimates from the relaxed MML problems are not perfectly consistent with each other. For example, $\widehat{\mbf{J}}^{\text{Relax}}_{i,j}$, which comes from node $i$'s local estimate, may not agree with $\widehat{\mbf{J}}^{\text{Relax}}_{j,i}$, which comes from node $j$'s local estimate. Therefore the resulting global estimate $\widehat{\mbf{J}}^{\text{Relax}}$ in \eqref{eq:estimatorRMMLE} is not guaranteed to be symmetric. 

A common way of addressing these discrepancies is to use iterative consensus methods as in~\cite{wiesel2012distributed, liu12a}. In this work however, we find that a single round of naive local averaging along edges is sufficient to ensure convergence to the true parameters, and also to yield a good approximation to the global MLE. Specifically, the local average is given by 
\ALGNN{
\label{eq:avg}
\widehat{\mbf{J}}_{i,j}^{\mathrm{Relax}} \leftarrow 
	\frac{1}{2}(\widehat{\mbf{J}}_{i,j}^{\mathrm{Relax}} + \widehat{\mbf{J}}_{j,i}^{\mathrm{Relax}}), \quad (i,j) \in E, 
}
which is the only message passing required. This message passing is single pass, unlike LBP which requires several iterations (if it converges at all).  
In the one-hop case, the resulting symmetric estimator coincides with the AVE estimator proposed in~\cite{wiesel2012distributed}.

\section{Analysis}
\label{sec:analysis}

\subsection{Asymptotic Analysis:  Classical Fixed-Dimensional 
Regime} 
\label{sec:asymptanalysis}

First we analyze the proposed distributed RMML estimator in the classical asymptotic regime, where the number of variables $p$ is fixed while the number of samples $T$ goes to infinity.
Let $\mbf{J}^{*}$ and $\Sigstar$ denote the true precision and covariance matrices, respectively. The following theorem states the asymptotic consistency of the RMML estimator $\widehat{\mbf{J}}^{\text{Relax}}$ and characterizes its asymptotic mean squared error:

\begin{theorem}[Asymptotic MSE]	\label{thm:asym}
The relaxed MML estimator $\widehat{\mbf{J}}^{\text{Relax}}$ is asymptotically consistent, and its mean squared (Frobenius) error satifies
\ALGNN{
{T} \cdot \mathbb{E} \| \widehat{\mbf{J}}^{\text{Relax}} - \mbf{J}^{*} \|_{F}^{2} \ \overset{T \rightarrow \infty}{\longrightarrow} \ \sum_{i=1}^{p} \sum_{j \in L_{i}} [\text{diag}\left( \mbf{F}_{i}^{-1} \right)]_{j},
}
where $T$ is the number of samples, $\text{diag}(\cdot)$ denotes the diagonal of a matrix, and $\mbf{F}_{i}$ is the Fisher information matrix of the relaxed MML problem in the $i${th} neighborhood~\eqref{eq:localrelax}, which takes the 
following form:
\ALGNN{
\left( \mbf{F}_{i} \right)_{(m,n), (l,k)} = 
\begin{cases}
2  {\mbf{\Sigma}^{*}_{m,l}}^{2}, & m=n \text{ and } l=k \\
2  \mbf{\Sigma}^{*}_{m,k}  \mbf{\Sigma}^{*}_{l,n}, & m=n, l \neq k \text{ or } m\neq n, l=k \\
\mbf{\Sigma}^{*}_{m,k}  \mbf{\Sigma}^{*}_{n,l}, & \text{otherwise}.
\end{cases}
\label{eq:fisher}
}
\end{theorem}

The above result can be derived by applying classical asymptotic theory~\cite{van2000asymptotic} to each local RMML problem~\eqref{eq:localrelax}, which is a well-defined M-estimation problem. Then the asymptotic behavior of the global RMML estimate follows by aggregation. 
The detailed proof of Theorem~\ref{thm:asym} is provided in Appendix~\ref{app:proofthm}.

While Theorem~\ref{thm:asym} ensures the consistency of RMML estimators with arbitrary local neighborhoods (as long as the row parameters are included), the following theorem guarantees that, in the asymptotic limit, larger neighborhoods always yield reduced estimation variance:

\begin{theorem}[Monotonicity of Asymptotic MSE]	\label{thm:variance}
Let $\widehat{\mbf{J}}^{\text{Relax, $k$-hop}}$ be the RMML estimate obtained from $k$-hop local neighborhoods. 
When the number of samples $T \rightarrow \infty$, for $k = 1, 2, \ldots$, we have
\ALGNN{
%\begin{split}
\mathbb{E} \| \widehat{\mbf{J}}^{\text{Relax, $k$-hop}} - \mbf{J}^{*} \|_{F}^{2} & \ge \mathbb{E} \| \widehat{\mbf{J}}^{\text{Relax, $(k+1)$-hop}} - \mbf{J}^{*} \|_{F}^{2} \label{eq:varinequality1}\\
& \ge \mathbb{E} \| \widehat{\mbf{J}}^{\text{GML}} - \mbf{J}^{*} \|_{F}^{2}. \label{eq:varinequality2}
%\end{split}	\label{eq:varinequality}
}
\end{theorem}

While Theorem~\ref{thm:variance} is stated for Gaussian graphical models, it was first  proven for the case of discrete graphical models by Massam and Wang in~\cite{massam2013distributed}.
%\footnote{The authors were made aware of~\cite{massam2013distributed} by an anonymous reviewer after submission of this manuscript.}. 
As pointed out by the authors of~\cite{massam2013distributed}, their proof can be easily extended to the Gaussian case. 
%However, for completeness, we include a self-contained proof of Theorem~\ref{thm:variance} in Appendix~\ref{sec:proof:var}.
%For completeness, in Appendix~\ref{sec:proof:var}, we include our own proof of Theorem~\ref{thm:variance} which actually parallels that given in~\cite{massam2013distributed}.
For completeness, we include our own proof of Theorem~\ref{thm:variance} in Appendix~\ref{sec:proof:var}.  The two proofs follow parallel lines of argument. 

%The proof of Theorem~\ref{thm:variance} is included in Appendix~\ref{sec:proof:var}.   
In Section~\ref{sec:exp}, we present numerical results that verify Theorem~\ref{thm:variance} not only in the large-sample regime but also when the sample size $T$ is comparable to or smaller than $p$.  In particular, it will be seen that the difference between $k = 1$ and $k = 2$ hops is most significant while the difference between $k = 2$ and the GML estimator (and by extension $k > 2$ and GML) is much smaller.

%%%%%%%%%%%%%%%%%%%%%%%%%%%%%%%%%%%%%%%%%%%
\subsection{Asymptotic Analysis: High-Dimensional Regime} % when $p,T \rightarrow \infty$}
\label{sec:nonasymp}

Theorems \ref{thm:asym} and \ref{thm:variance} characterize the classical asymptotic behavior of the RMML estimator.
In this subsection we analyze the high-dimensional convergence rate of the RMML estimator, which can be applied to settings 
where both the number of variables $p$ and the number of samples $T$ increase to infinity, i.e. $p,T \rightarrow \infty$. Such problems arise in high-dimensional applications, and have attracted much attention in modern statistics~\cite{ravikumar2011high, friedman2008sparse, rothman2008sparse}.
We will show that under very mild conditions,
the proposed RMML estimator enjoys a sharp MSE convergence rate to the true parameter, which is almost the same as the more expensive global ML estimator.

Similar to~\cite{rothman2008sparse, ravikumar2011high}, we first assume that the maximum eigenvalue of $\Jstar$ is bounded from above:
\ALGNN{
\la_{\max}(\Jstar) \le \overline{\kappa} < \infty.
\label{eq:upperboundeigen}
}
Recall that $R_{i}$ defines the relaxed edge set in the $i^{th}$ local neighborhood. Let 
$\overline{R}$ denote the maximum cardinality among all local relaxed edge sets, i.e.
\ALGNN{
\overline{R} := \max_{i = 1,\ldots, p} |R_{i}|,
}
and let 
$r$ denote the sum of the cardinalities of all local relaxed edge sets:
\ALGNN{
 r := \sum_{i = 1}^p | R_i |.
 }
Also denote 
$\overline{\sigma} := \max_{i=1,\ldots,p} \Sigstar_{i,i}$ as the maximum variance.

The following theorem states an upper bound on the estimation error rate in the high-dimensional regime.
\begin{theorem}[High-dimensional MSE]	\label{thm:nonasym}
Assume the number of samples $T$ satisfies 
\ALGNN{
T \ge {C^{2} c_1 \log p},
\label{eq:samplecomplexity1}
}
for $c_1 = {6400  \overline{\sigma}^{2}}/{\min^{2} \{ \frac{1}{9 \overline{\kappa} \sqrt{\overline{R}}}, 40 \overline{\sigma} \}}$ and an arbitrary constant $C \ge 1$. Then
\ALGNN{
\| \JRelax - \Jstar \|_F 
\le 720 C \cdot \overline{\kappa}^2  \overline{\sigma}  \sqrt{\frac{r \log p }{T}},
}
with probability greater than $1 - {4}/{ p^{2(C^2 - 1)}}$. 
\end{theorem}

Proof of 
Theorem~\ref{thm:nonasym} can be found in Appendix~\ref{sec:proof:nonasym}.

\vspace{1em}
\noindent \emph{Remarks:} \\
1) It is interesting to compare the result in Theorem~\ref{thm:nonasym} with the standard convergence rate for the GML estimator (e.g., \cite{rothman2008sparse, ravikumar2011high, wainwright2009sharp}). Theorem~\ref{thm:nonasym} assumes a very mild condition (Eq.~\eqref{eq:samplecomplexity1}) on the sample size, which is less restrictive than the requirement $O(p \log p)$ shown in~\cite{rothman2008sparse} in the high dimensional regime, and is comparable to those obtained in~\cite{ravikumar2011high, wainwright2009sharp} when the local neighborhood size increases more slowly than $p$, i.e. $\overline{R} = o(p)$. However, we emphasize that, unlike some of the literature, we assume the graph structure is known. \\
2) The error bound in Theorem~\ref{thm:nonasym} is (up to a constant) slightly more pessimistic than the rate $O(\sqrt{p \log p /T})$ shown in~\cite{rothman2008sparse, ravikumar2011high} by the additional factor of $r/p = \frac{\sum_{i=1}^p |R_i |}{p}$, which is roughly the average cardinality of local neighborhoods. Again, when the local neighborhood size increases more slowly than $p$ in the high-dimensional regime, this additional factor becomes relatively insignificant. \\
3) The mild sample size requirement is partly due to our distributed framework, under which the stochastic deviation is smaller since a smaller set of parameters needs to be considered for each local RMML problem. However, the additional parameters  introduced by convex relaxation and the aggregation of local estimation errors result in the additional factor $r/p$ mentioned above. This  demonstrates the trade-off   due to the desire for distributed, convex optimization in the proposed framework. 

%%%%%%%%%%%%%%%%%%%%%%%%%%%%%%%%%%%%%%%%%%%
\subsection{Robustness Against Model Mismatch}
\label{sec:robust}

One of the premises of the estimation framework we consider in this paper is that the true structure of the graph is known. However, this assumption could be violated in practice. In this section, we investigate the robustness of the estimators against small structure mismatch.   
Our specific interest is in the bias due to model mismatch and hence we focus on the infinite sample regime.

We first consider the GML problem. The GML estimator effectively provides a mapping from the edge elements of moment (covariance) parameters $\Sighat_{\Etilde}$ to the canonical (concentration) parameters $\widehat{\mbf{J}}^{\text{GML}}_{\Etilde}$. We denote this mapping as $\mcal{M}(\cdot; \Etilde)$, i.e., $\widehat{\mbf{J}}^{\text{GML}}_{\Etilde} = \mcal{M}(\Sighat; \Etilde)$. 
This mapping is specified implicitly by the optimality condition:
\ALGNN{
{\Sighat}_{\Etilde} - \left(\left( \widehat{\mbf{J}}^{\text{GML}} \right)^{-1}\right)_{\Etilde} = \mbf{0}.	\label{eq:optimalcond}
}
Due to a property of minimal exponential families, $\mcal{M}(\cdot; \Etilde)$ exists and is unique provided that covariance matrix $\Sighat$ is positive definite \cite{wainwright2008graphical}. 
Also by the implicit function theorem, $\mcal{M}(\cdot; \Etilde)$ 
is differentiable and thus continuous. 

Consider a perturbed concentration matrix $\widetilde{\mbf{J}}^{*}$ which has   uniformly bounded perturbations on the non-edge entries with respect to the nominal parameter $\Jstar$:
\ALGNN{
\widetilde{\mbf{J}}^{*} = \mbf{J}^{*} + \Delta {\mbf{J}},	\label{eq:perturbmodel}
}
where $\Delta {\mbf{J}}$   is supported only on $\Etilde^C$. We assume the perturbation is small enough, such that the perturbed matrix is still positive definite. Denote the corresponding covariance matrix as $\widetilde{\mbf{\Sigma}} = (\widetilde{\mbf{J}}^{*})^{-1}$.
Then the bias of the GML estimator due to model perturbation can be obtained by a first-order perturbation analysis of the GML mapping defined above.

Let $\Gamma = \mbf{\Sigma} \otimes \mbf{\Sigma}$ denote the Hessian of the GML problem~\eqref{eq:gml} with no sparsity constraints, 
which is also related to the Jacobian of~\eqref{eq:optimalcond} with respect to $\widehat{\mbf{J}}^{\text{GML}}_{\Etilde}$. We have
\ALGN{
\widetilde{\mbf{J}}^{\text{GML}}_{\Etilde} & = \mcal{M}(\widetilde{\mbf{\Sigma}}; \Etilde) \\
& = \mcal{M}((\widetilde{\mbf{J}}^{*})^{-1}; \Etilde) \\
& = \mcal{M}(( \mbf{J}^{*} + \Delta {\mbf{J}} )^{-1}; \Etilde) \\
& = \mcal{M}({\mbf{J}^{*}}^{-1} + \Gamma_{\Etilde, \Etilde^{C}}\Delta \mbf{J}_{\Etilde^{C}} + O(\| \Delta \mbf{J} \|^2); \Etilde) \\
& = \mcal{M}(\mbf{\Sigma}^{*}; \Etilde) + (\Gamma_{\Etilde, \Etilde})^{-1}\Gamma_{\Etilde, \Etilde^{C}}\Delta \mbf{J}_{\Etilde^{C}} + O(\| \Delta \mbf{J} \|^2),
}
where in the second-to-last relation we have used the first-order approximation of matrix inversion, and the last identity is due to the implicit function theorem applied to the optimality condition~\eqref{eq:optimalcond}. Also note that $\mbf{J}^* = \mcal{M}(\mbf{\Sigma}^{*}; \Etilde)$ due to consistency of the GML estimator. 

Therefore the maximum element-wise bias with respect to the new model can be bounded as follows (disregarding higher-order terms):
\ALGNN{
\begin{split}
& \| \widetilde{\mbf{J}}^{\text{GML}} - \widetilde{\mbf{J}}^{*} \|_{\infty}  \\
& \le \| \widetilde{\mbf{J}}^{\text{GML}} - \mbf{J}^{*} \|_\infty + \|  \widetilde{\mbf{J}}^{*} - \mbf{J}^{*} \|_\infty \\
& \lesssim  \vertiii{ (\Gamma_{\Etilde,\Etilde})^{-1}\Gamma_{\Etilde,\Etilde^{C}}}_{\infty}   \| \Delta \mbf{J}_{\Etilde^{C}} \|_{\infty} +  \| \Delta \mbf{J}_{\Etilde^{C}} \|_{\infty},
\end{split}
\label{eq:robustbound}
}
where we recall $\vertiii{\cdot}$ is the induced $\infty/\infty$ matrix norm.
%------------------------

The second term in the last display is the inevitable 
bias due to model mismatch, 
while the first term captures the additional bias 
attributable to the GML estimator under model perturbation. 
The additional bias depends on $\vertiii{(\Gamma_{\Etilde,\Etilde})^{-1}\Gamma_{\Etilde,\Etilde^{C}}}_{\infty}$, which is intuitively related to the level of \emph{incoherence} between the edge and non-edge elements in the Hessian of the GML problem~\eqref{eq:gml}. Similar incoherence quantities have been shown to play a crucial role in the literature on variable selection~\cite{meinshausen2006high} (e.g. Lasso) and structure estimation in Gaussian graphical models~\cite{ravikumar2011high}. Therefore the smaller this incoherence parameter is, the more robust the GML estimator will be.

Since each local problem in RMML estimation has the same structure as the GML problem, we can apply similar analysis to each local neighborhood. The resulting bound on the bias of the RMML estimator is dependent on  similar incoherence parameters but defined with respect to   relaxed edge sets in the local neighborhoods. We conjecture that these local incoherence parameters are comparable to, if not smaller than, the global incoherence. Hence the robustness of the distributed RMML estimator is expected to be comparable to the GML estimator.   While our conjecture is not formally proven in this paper, it is positively supported by the numerical experiments in Sec.~\ref{sec:exp}.

\section{Computational Complexity and Implementation} 
\label{sec:computation}

In this section we discuss the computational complexity of the proposed RMML approach and some implementation issues. First we note that each local RMML problem has the same structure as the centralized ML problem, which is a $\log\det$-regularized semidefinite program ($\log\det$-SDP). Therefore many well-developed solvers and efficient specialized algorithms can be used. 
Furthermore, due to the distributed nature of the RMML approach, the local problems can all be solved in parallel before the final one-step averaging. 
  The combination of lower dimensionality in the local problems and parallelization can significantly reduce the total run time compared with centralized algorithms.

  In terms of algorithms,  we find the iterative regression method introduced in~\cite{friedman2009elements} is very efficient for sparse graphs. This algorithm iteratively performs linear regressions of each node variable against its immediate neighbors until global convergence. However, the major drawback of this algorithm is the need to maintain global parameters, which prevents direct parallelization and also   makes implementation difficult  in distributed networks (as discussed below).

The   computational advantage of the proposed RMML algorithm becomes more obvious when the number of variables $p$ increases to large numbers. 
Assuming that the local neighborhood dimensions increase more slowly than $p$, such as with K-NN graphs and lattice graphs, the total complexity of the RMML estimator scales linearly in $p$, independent of the algorithm used  to solve the local problems. The run time increases even   more slowly if the overall algorithm can be parallelized. 
  In contrast, for the centralized algorithms, the dependence of complexity on $p$ is at least linear and is much faster for denser graphs and/or if generic $\log\det$-SDP solvers are used. 

Another advantage of the proposed RMML algorithm is that it is highly suitable for network applications due to its minimal requirement for message passing which reduces  communication cost. 
In sharp contrast, many centralized algorithms, such as the iterative regression algorithm mentioned above,
require centralized storage and iterative updating of a large number of variables, which in turn requires expensive communication among non-adjacent nodes in the network.

\begin{figure}[t]
\centering
\includegraphics[width=0.4\textwidth]{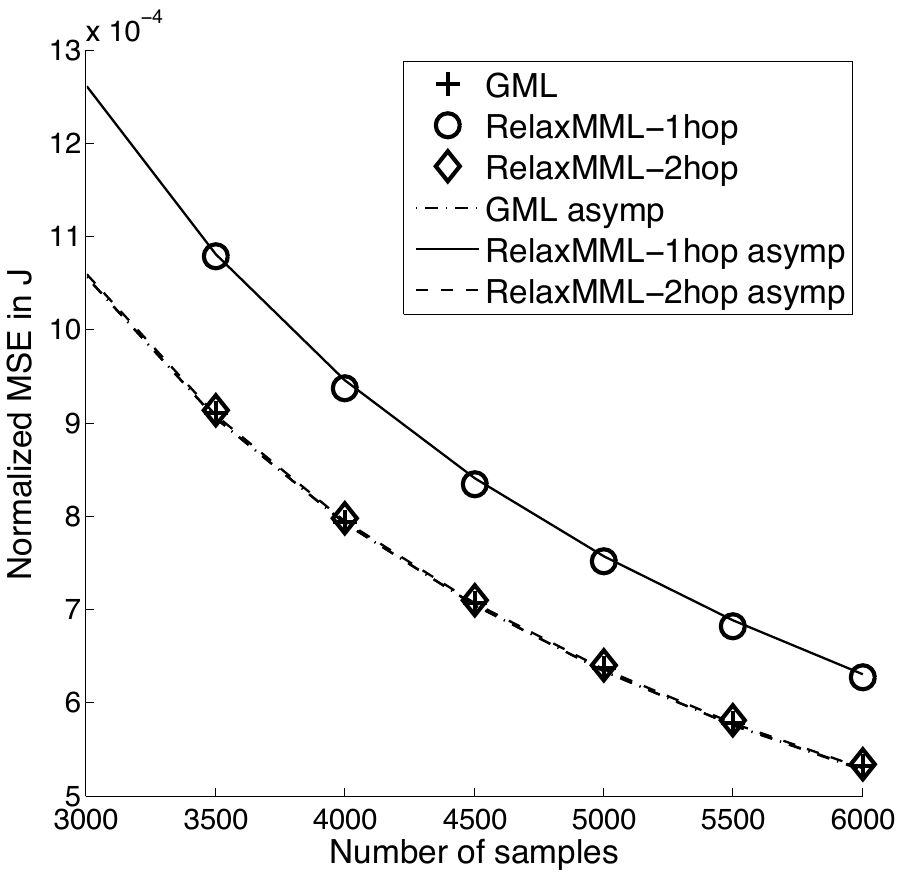}
\caption{Asymptotic normalized MSE for K-NN graphs ($p = 20, K = 4$).
The curves denote the theoretical asymptotic limits, whereas the symbols denote the empirical normalized MSE over 10,000 runs. 
}
\label{subfig:asympvar}
\end{figure}
%%%%%%%%%%%%%%%%%%%%%%%%%%%%%%%%%%%%%%%%

\begin{figure*}[t!]
\begin{center}
	\begin{subfigure}[Normalized MSE for K-NN graphs ($p = 500, K = 4$) ]{
	\label{subfig:mseKNN}
		\centering
		\includegraphics[width=0.4\textwidth]{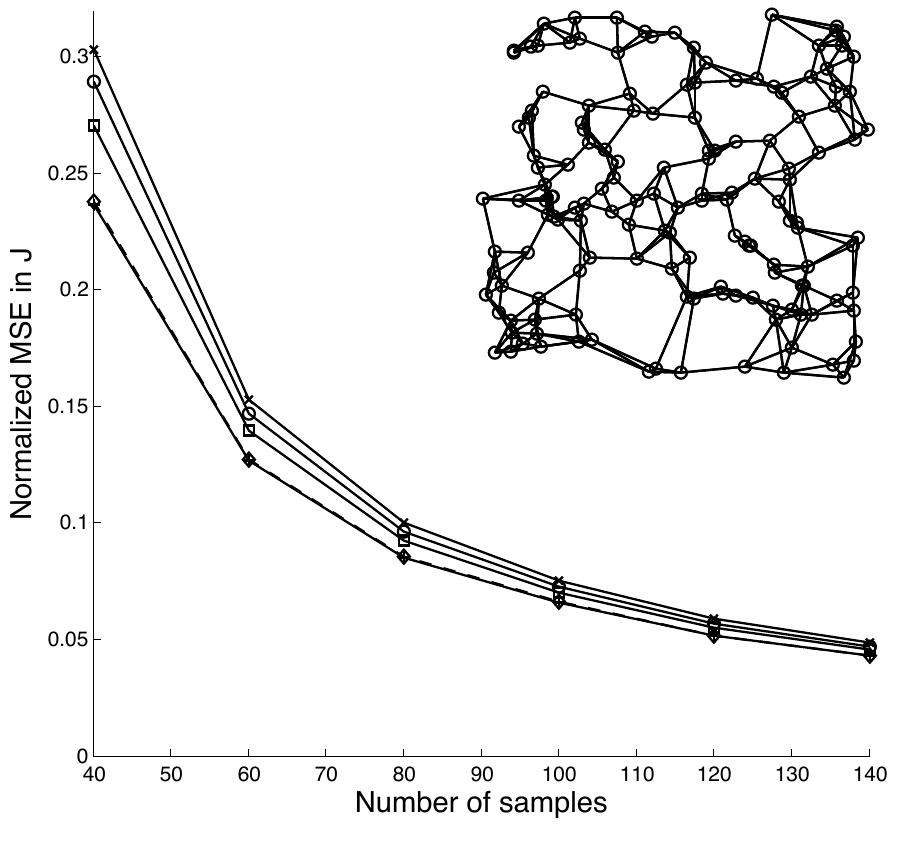}
	}
	\end{subfigure}
%	~
	\begin{subfigure}[Normalized MSE for lattice graphs ($p=20\times20=400$, $\mu = 0.5$, $\sigma^{2} = 0.2$)]{
	\label{subfig:mseGrid}
		\centering
		\includegraphics[width=0.4\textwidth]{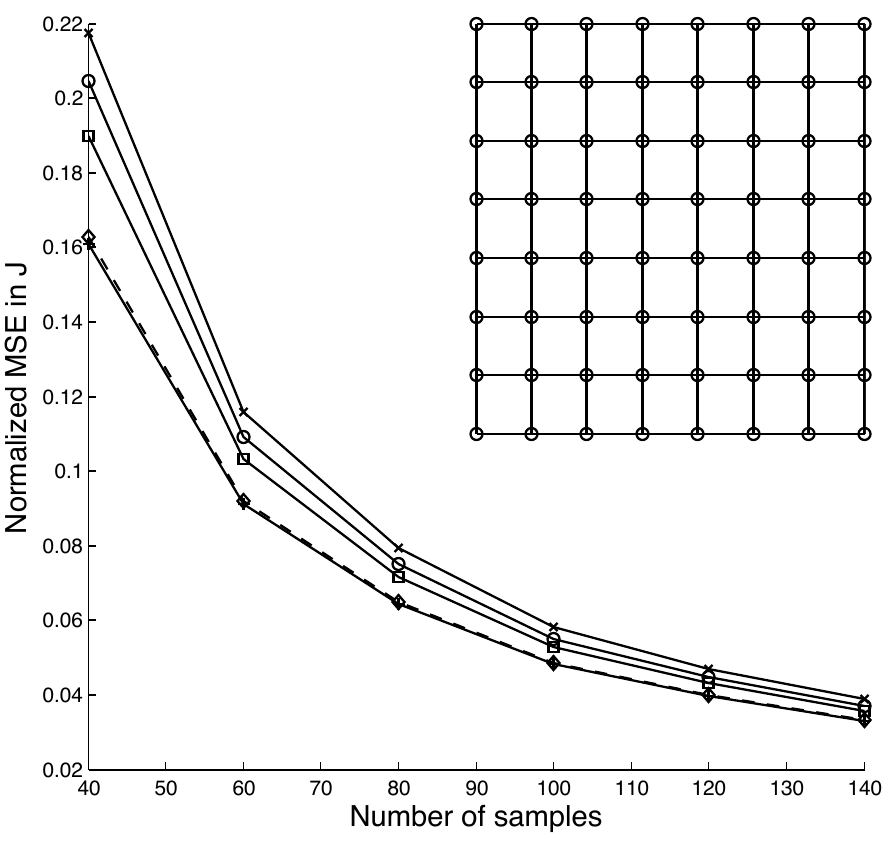}
	}
	\end{subfigure}
%	\\
	\begin{subfigure}[Normalized MSE for small-world graphs ($p = 100$, $K = 20$, $\beta = 0.5$)]{
	\label{subfig:mseSmallworld}
		\centering
		\includegraphics[width=0.4\textwidth]{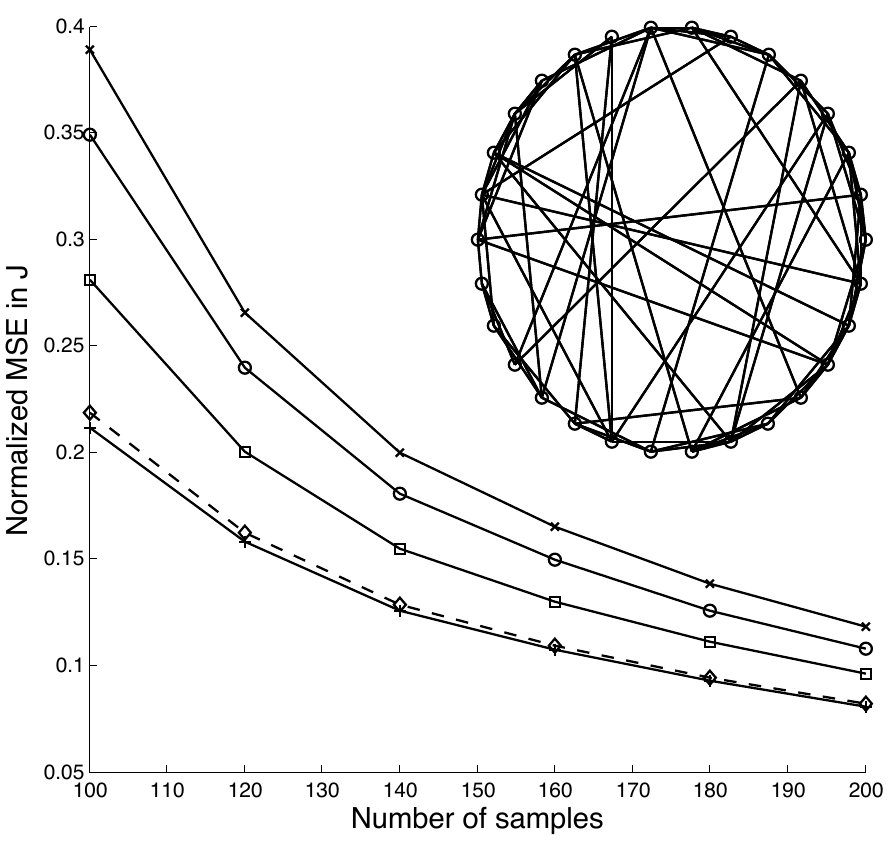}
	}
	\end{subfigure}
%	~ 
	\begin{subfigure}[Normalized MSE for IntelLab Sensor Network Data set~\cite{guestrin2004distributed} ($p = 50$)]{
	\label{subfig:mseWSN}
		\centering
		\includegraphics[width=0.4\textwidth]{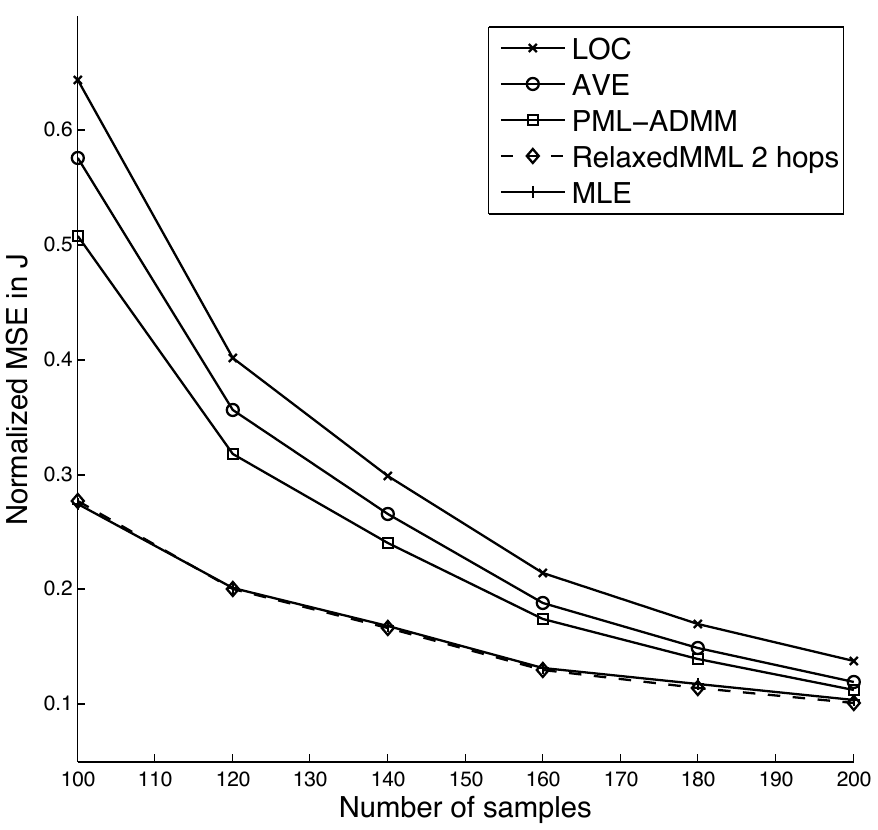}
	}
	\end{subfigure}
\caption{Normalized MSE in the concentration matrix estimates for different graphical models. The legend in Figure~\ref{subfig:mseWSN} applies to all plots. The proposed 2-hop relaxed maximum marginal likelihood (RMML) estimator clearly improves upon existing distributed estimators and nearly closes the gap to the centralized maximum likelihood estimator.
}
\label{fig:exp}
\end{center}
\end{figure*}

\section{Experiments}
\label{sec:exp}

In this section, we evaluate the proposed RMML estimator and compare it with the centralized and other distributed estimators in the literature. All methods have been coded in Matlab routines that will be available at the reproducible research web page~\footnote{\url{http://tbayes.eecs.umich.edu/rrpapers}}. We focus on the one-hop and two-hop versions of the RMML estimator (denoted as 
\texttt{RelaxMML-1hop} and \texttt{RelaxMML-2hop}, respectively). Other estimators considered in this section are:
	\begin{itemize}
	\item The centralized GML estimator, denoted as \texttt{GML} in the legends; 
	\item The \emph{LOCAL} and \emph{AVE} estimators from~\cite{wiesel2012distributed}, denoted as \texttt{LOC} and $\texttt{AVE}$. They coincide with the asymmetric and symmetric versions respectively of the \emph{one-hop} relaxed MML estimator;
	\item The weighted maximum pseudo-likelihood estimator using Alternating Direction Method of Multipliers (ADMM) consensus, proposed in~\cite{wiesel2012distributed} and~\cite{liu12a} and denoted as \texttt{PML-ADMM}.  We use the weights $\left[ \hat{\mbf{J}}_{i,i}^{LOC} \right]^{2}$ as in~\cite{wiesel2012distributed}. 
	\end{itemize}

We first verify the classical asymptotic rates for the proposed estimators predicted by Theorems~\ref{thm:asym} and~\ref{thm:variance} (see Fig.~\ref{subfig:asympvar}) using 10,000 randomized runs sampled from a four-nearest-neighbor Gaussian graphical model with $p = 20$ nodes  distributed uniformly  in space over the unit square. 
The concentration matrix is initialized as $\mbf{J}_{i,j} = \pm \exp(-0.5\cdot d_{i,j})$ with random sign, where $d_{i,j}$ is the Euclidean distance between the $i$th and $j$th nodes. 
%Finally we add a small value to the diagonal to ensure positive definiteness. 	
The empirical normalized mean squared errors (MSE), defined as 
%\ALGN{
$\frac{\| \widehat{\mbf{J}} - \mbf{J} \|^{2}_{F}}{\| \mbf{J} \|^{2}_{F}}$,
%}
%$\frac{\| \widehat{\mbf{J}} - \mbf{J} \|^{2}_{F}}{\| \mbf{J} \|^{2}_{F}}$ 
are computed from Monte Carlo samples, and they are compared with the theoretical bounds predicted by Theorem~\ref{thm:asym}. Fig.~\ref{subfig:asympvar}  illustrates the tightness of these bounds. It is also worth noting that the bound for the two-hop RMML estimator is much lower than that of the one-hop estimator, as predicted by Theorem~\ref{thm:variance}. The two-hop bound approximates the bound for the GML estimator closely, suggesting that RMML estimators are nearly asymptotically efficient. The asymptotic bounds for RMML estimators with larger neighborhoods follow the monotonicity relation in Theorem~\ref{thm:variance}, however the differences are too small to visually identify, and hence are omitted from the plot.

Next we evaluate the non-asymptotic MSE performance of the proposed estimator, and compare it with the other estimators on both synthetic and real-world data sets. For synthetic data sets, 
%In our experiments, 
we consider three classes of graphs that are motivated by real-world applications. For each class we follow similar experiment settings as in~\cite{wiesel2012distributed}. Specifically, we randomly generate $20$ topologies and associated sparse concentration matrices $\mbf{J}$, and for each $\mbf{J}$, we perform 10 experiments in which random samples are drawn from the distribution and the concentration matrix is estimated from the samples. The normalized MSEs are 
 averaged over all $200$ experiments, and are reported in Figure~\ref{fig:exp}. 
 An illustration of the graph topology is shown in the top-right corner of each plot. 
 The classes of graphs we consider are:
\begin{itemize}
	\item \textbf{K-NN graphs} (Figure~\ref{subfig:mseKNN}): A K-nearest neighbor graph is a straightforward model for real-world networks whose measurements have correlations that depend on pairwise Euclidean distances, \emph{e.g.}, sensor networks. For these experiments, we randomly generate $p = 500$ nodes uniformly over the unit square. Each node is then connected to its $K$-nearest neighbors, where $K = 4$.  The concentration matrix is initialized as $\mbf{J}_{i,j} = \pm \exp(-0.5\cdot d_{i,j})$ with random sign, where $d_{i,j}$ is the Euclidean distance between the $i$th and $j$th nodes. Finally we add a small value to the diagonal to ensure positive definiteness. 	
	\item \textbf{Lattice graphs} (Figure~\ref{subfig:mseGrid}): A lattice graph is appropriate for networks with regular spatial correlations, e.g., images that are Markov random fields. We generate a square lattice graph with $p = 20 \times 20 = 400$ nodes and edge weights generated as $\mbf{J}_{i.j} = \min\{ w, 1 \}$, where $w$ is a normally distributed random variable with mean 0.5 and variance 0.2. A small value is added to the diagonal to ensure positive definiteness.	
	\item \textbf{Small-world graphs} (Figure~\ref{subfig:mseSmallworld}): Small-world graphs have been proposed for social networks, biological networks, etc., where most nodes have few immediate neighbors but can be reached from any other node through a small number of hops~\cite{watts1998small}. We generate graphical models structured as random small-world networks using the Watts-Strogatz mechanism~\cite{watts1998small} with $p = 100$, $K \text{(mean degree)} = 20$, and parameter $\beta = 0.5$. Under this particular setting, a large fraction of nodes have large second-hop neighborhoods with dimension close to $p$. In general we expect the second-hop neighborhood to scale linearly with respect to $p$. We choose the edge weights to be uniformly distributed and also add a small diagonal loading to ensure that $\mbf{J}$ is positive definite.
	\end{itemize}

%%%%%%%%%%%%%%%%%%%%%%%%%%%%%%%%%%%%%%%

\begin{figure}[th]
\begin{center}
\includegraphics[width=0.4\textwidth]{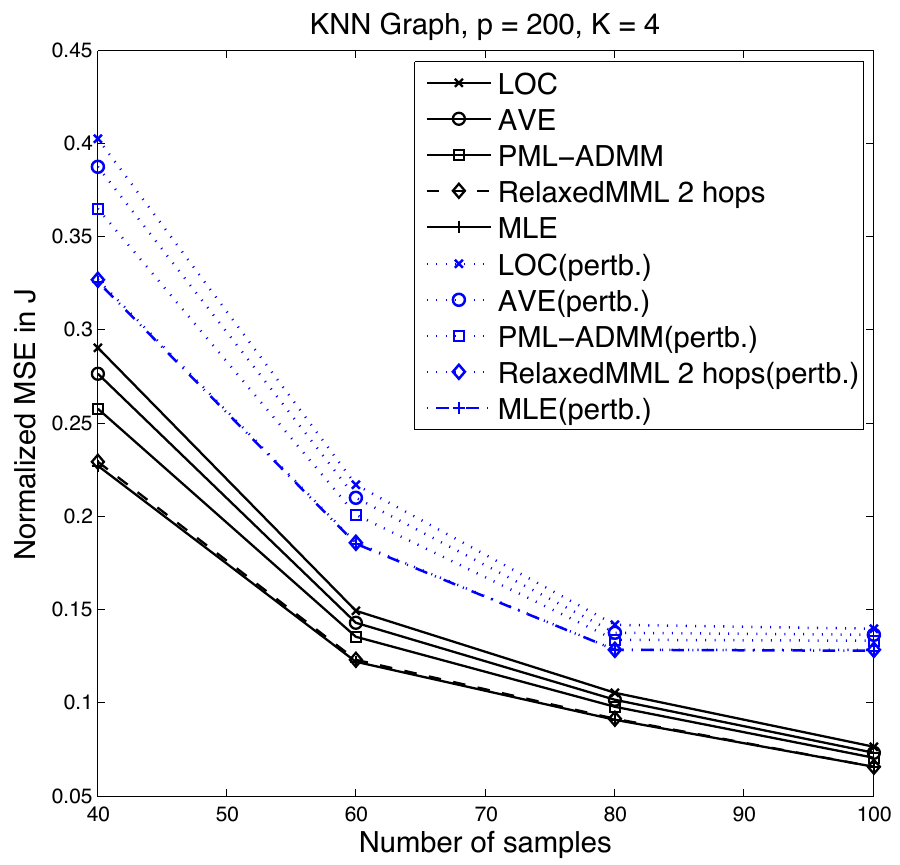}
\caption{Robustness of estimators under model mismatch. All errors are obtained from K-NN ($p=200, K=4$) graphs and averaged over 50 experiments. For the perturbed models, $\pm 0.1$ is added to the non-edge components of the nominal precision matrix. The proposed distributed RMML estimator is as robust as the GML estimator.}
\label{fig:robust}
\end{center}
\end{figure}

%%%%%%%%%%%%%%%%%%%%%%%%%%%%%%%%%%%%%%%

%%%%%%%%%%%%%%%%%%%%%%%%%%%%%%%%%%%%%%%%%%%%%%%%%%
\begin{figure*}[t!]
\begin{center}
	\begin{subfigure}[Run time comparison using \texttt{logdetPPA} solver]{
	\label{subfig:runtimeLogdet}
		\centering
		\includegraphics[width=0.4\textwidth]{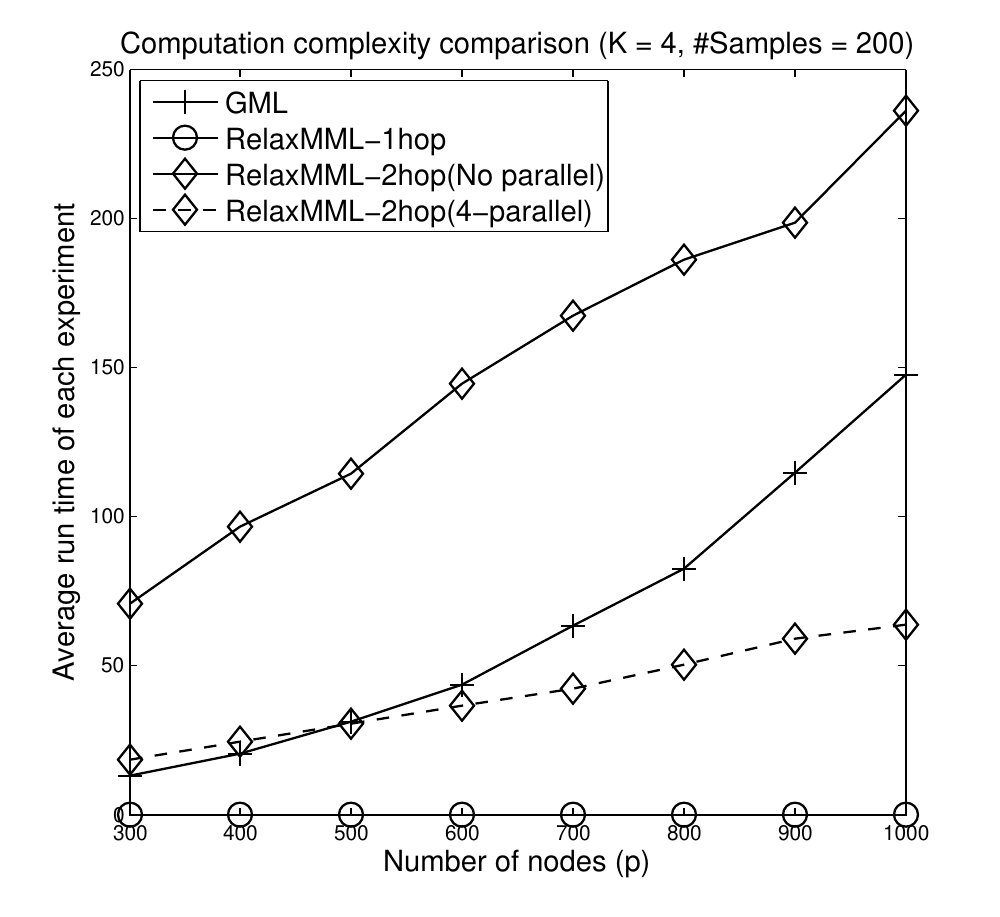}
	}
	\end{subfigure}
%	~
	\begin{subfigure}[Run time comparison using iterative regression algorithm]{
	\label{subfig:runtimeGlasso}
		\centering
        \includegraphics[width=0.4\textwidth]{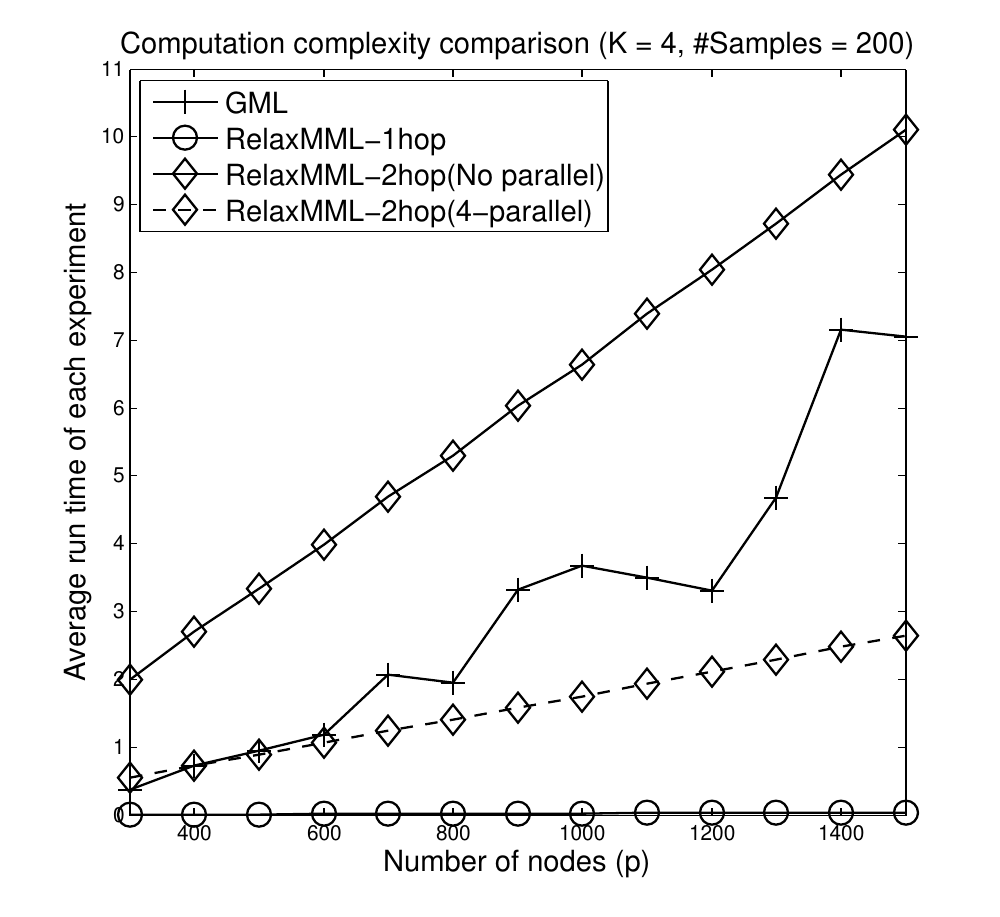}
	}
	\end{subfigure}
\caption{Run time comparisons for GML and RMML estimators. In panel (a) the \texttt{logdetPPA} solver is used, and in panel (b) the iterative regression algorithm is used. In both figures, solid lines denote the runtime scaling of the sequential version of the algorithm, while the dashed lines denote runtime scaling for a parallelized version with four cores. In both figures, the runtime of the GML estimator is super-linear in $p$, while the RMML estimator exhibits linear scaling in $p$, and the runtime is further reduced by a factor approximately equal to the number of cores used. All experiments are implemented in a Matlab environment.}
\label{fig:runTime}
\end{center}
\end{figure*}
%%%%%%%%%%%%%%%%%%%%%%%%%%%%%%%%%%%%%%%%%%%%%%%%%%

The MSE curves shown in Figure~\ref{fig:exp} match our theoretical predictions in Section~\ref{sec:nonasymp}, and they also demonstrate the superior performance of the proposed RMML estimator. In particular, for the graphs that have relatively small two-hop neighborhoods, namely the K-NN graphs and the lattice grids, the MSE of the proposed two-hop relaxed MML estimator almost coincides with the MSE of the global MLE. 
On the other hand, for small-world networks, the dimensions of the two-hop neighborhoods grow as fast as $p$. In this case, a noticeable gap emerges between the global MLE and the  two-hop relaxed MML estimator. 
These graphs are known to be harder to learn through distributed algorithms. 
The two-hop relaxed MML estimator still outperforms the other distributed algorithms by a large margin.

Next, we apply the estimators to a real-world sensor network.  
The {IntelLab} dataset~\cite{guestrin2004distributed} contains temperature information from a sensor network of 54 nodes deployed in the Intel Berkeley Research lab between February 28 and April 5, 2004. This dataset is known to be very difficult with missing data, noise and failed sensors. We select 50 sensors with relatively stable and regular measurements. To obtain a target concentration matrix, we use 1800 consecutive samples per sensor, interpolate the missing or failed readings and de-trend the data using a local rectangular window of 10 samples. Next, we compute the sample covariance and invert it to obtain a sample concentration matrix. This concentration matrix is then thresholded to yield a ground truth graphical model with a sparsity level of 70\% zeros. Using knowledge of the sparsity and sampling from the original 1800 samples, we estimate the concentration matrix using the same estimators as before. As shown in Figure~\ref{subfig:mseWSN}, the proposed two-hop relaxed MML estimator still gives a very tight approximation to the centralized GML estimator and its advantage over other distributed estimators is obvious.

We  investigate the robustness of the centralized and distributed estimators in the presence of model mismatch. The nominal precision matrix $\mbf{J}$ corresponds to a four-nearest-neighbor graphical model with $p=200$ as in the previous experiments. We add $\pm 0.1$ random perturbations to the non-edge components of the nominal precision matrix (also with minimal diagonal loading to ensure positive definiteness), then generate samples from the perturbed model. The different estimation algorithms are applied assuming the nominal graph structure and the resulting MSEs are plotted with respect to the nominal model. The MSEs of all estimators (using samples from both the original and perturbed models, respectively) are reported in Figure~\ref{fig:robust}. All errors are averaged across 50 randomized experiments. As can be seen, the model mismatch leads to estimation bias for both centralized and distributed estimators. The magnitudes of the model mismatch bias for all estimators are comparable, as predicted by the theoretical analysis in Sec.~\ref{sec:robust}. These experiment results confirm the robustness of the proposed distributed algorithm.

We next turn to computational comparisons. In the following experiments, we illustrate the computational gain of our distributed estimator over the centralized one through two runtime comparisons performed in Matlab. Our main focus is on the relative scaling of the runtime 
with respect to the number of nodes $p$ for different estimators. 
We consider two algorithms for solving both the centralized GML problem and the local RMML problems. The first is an interior point algorithm implemented in the solver \texttt{logdetPPA}~\cite{wang2010solving}, which is specially designed for solving $\log \det$-SDPs. The second algorithm is the iterative regression approach in~\cite{friedman2009elements} for solving the covariance selection problem~\cite{friedman2008sparse} with known structure. In both experiments, the graphical model is a four-NN graph with similar parameter settings as before. We compare the total runtime of the GML estimator and that of different versions of RMML estimators. For the RMML estimators, we implement a sequential and a parallel version using the \texttt{parfor} function in {Matlab}. The results are reported in Figure~\ref{fig:runTime}. As expected, the runtime of the GML estimator is at least linear in $p$ and the generic solver appears to be much more expensive than the iterative regression algorithm for this particular task. The total cost of the RMML estimator without parallelization is also linear in $p$, and is slightly higher than the GML estimator. However, when four-core parallelization is used, the run time is approximately reduced by a factor of four, resulting in lower computational complexity after $p > 500$. 

It is expected that with a higher degree of parallelization, the run time of the proposed RMML estimator will continue to decrease almost linearly with the number of cores.  As discussed in Section~\ref{sec:computation}, all local RMML problems can be solved in parallel without the need for any iterative message-passing.  Therefore the communication overhead is minimal, consisting of the final concatenation and symmetrization steps~\eqref{eq:estimatorRMMLE} and~\eqref{eq:avg}.

\section{Conclusion}
\label{sec:conclusion}

We have proposed a distributed MML framework for estimating the concentration matrix in Gaussian graphical models. The proposed method solves a convex relaxation of a marginal likelihood maximization problem independently in each local neighborhood. A global estimate is then obtained by combining the local estimates via a single round of local averaging. The proposed estimator is shown to be statistically consistent and computationally efficient. 
In particular, we have shown that the statistical convergence rate of our estimator is comparable to that of the more expensive centralized maximum likelihood estimator. Likewise in numerical experiments, a two-hop version of the distributed estimator is seen to be sufficient to attain centralized performance.
Its improved performance relative to existing distributed estimators is also illustrated.

%There are several directions that are worthy of further study.
%First, it would be worthwhile to investigate the extension of the proposed framework to Markov random fields with non-Gaussian or discrete distributions (see~\cite{mizrahi2013efficient, massam2013distributed} for recent efforts along this direction). In addition to parameter estimation, another interesting direction is to also estimate the structure of graphical models in a similar distributed fashion. This appears to be a harder problem, but some insights regarding distributed algorithms and convex relaxation in particular can perhaps be shared. 

\bibliographystyle{IEEEtran}
\bibliography{ref-tsp}

\begin{thebibliography}{10}
\providecommand{\url}[1]{#1}
\csname url@samestyle\endcsname
\providecommand{\newblock}{\relax}
\providecommand{\bibinfo}[2]{#2}
\providecommand{\BIBentrySTDinterwordspacing}{\spaceskip=0pt\relax}
\providecommand{\BIBentryALTinterwordstretchfactor}{4}
\providecommand{\BIBentryALTinterwordspacing}{\spaceskip=\fontdimen2\font plus
\BIBentryALTinterwordstretchfactor\fontdimen3\font minus
  \fontdimen4\font\relax}
\providecommand{\BIBforeignlanguage}[2]{{%
\expandafter\ifx\csname l@#1\endcsname\relax
\typeout{** WARNING: IEEEtran.bst: No hyphenation pattern has been}%
\typeout{** loaded for the language `#1'. Using the pattern for}%
\typeout{** the default language instead.}%
\else
\language=\csname l@#1\endcsname
\fi
#2}}
\providecommand{\BIBdecl}{\relax}
\BIBdecl

\bibitem{meng2013distributed}
Z.~Meng, D.~Wei, A.~Hero, and A.~Wiesel, ``Distributed learning of {Gaussian}
  graphical models via marginal likelihoods,'' in \emph{Proceedings of the
  Sixteenth International Conference on Artificial Intelligence and
  Statistics}, 2013, pp. 39--47.

\bibitem{meng2013marginal}
------, ``Marginal likelihoods for distributed estimation of graphical model
  parameters,'' in \emph{Computational Advances in Multi-Sensor Adaptive
  Processing (CAMSAP), 2013 IEEE 5th International Workshop on}, Dec 2013, pp.
  73--76.

\bibitem{lauritzen1996graphical}
S.~Lauritzen, \emph{Graphical models}.\hskip 1em plus 0.5em minus 0.4em\relax
  Oxford University Press, USA, 1996, vol.~17.

\bibitem{wainwright2008graphical}
M.~Wainwright and M.~Jordan, ``Graphical models, exponential families, and
  variational inference,'' \emph{Foundations and Trends{\textregistered} in
  Machine Learning}, vol.~1, no. 1-2, pp. 1--305, 2008.

\bibitem{liu12a}
Q.~Liu and A.~Ihler, ``Distributed parameter estimation via
  pseudo-likelihood,'' \emph{International Conference on Machine Learning
  (ICML)}, Jun. 2012.

\bibitem{wiesel2012distributed}
A.~Wiesel and A.~Hero, ``Distributed covariance estimation in {Gaussian}
  graphical models,'' \emph{Signal Processing, IEEE Transactions on}, vol.~60,
  no.~1, pp. 211--220, 2012.

\bibitem{meng2012distributed}
Z.~Meng, A.~Wiesel, and A.~Hero, ``Distributed principal component analysis on
  networks via directed graphical models,'' in \emph{Acoustics, Speech and
  Signal Processing (ICASSP), 2012 IEEE International Conference on}.\hskip 1em
  plus 0.5em minus 0.4em\relax IEEE, 2012, pp. 2877--2880.

\bibitem{banerjee2006convex}
O.~Banerjee, L.~El~Ghaoui, A.~d'Aspremont, and G.~Natsoulis, ``Convex
  optimization techniques for fitting sparse {Gaussian} graphical models,'' in
  \emph{ACM International Conference Proceeding Series}, vol. 148.\hskip 1em
  plus 0.5em minus 0.4em\relax Citeseer, 2006, pp. 89--96.

\bibitem{dahl2008covariance}
J.~Dahl, L.~Vandenberghe, and V.~Roychowdhury, ``Covariance selection for
  non-chordal graphs via chordal embedding,'' \emph{Optimization Methods and
  Software}, vol. 23(4), pp. 501--520, 2008.

\bibitem{friedman2008sparse}
J.~Friedman, T.~Hastie, and R.~Tibshirani, ``Sparse inverse covariance
  estimation with the graphical lasso,'' \emph{Biostatistics}, vol.~9, no.~3,
  pp. 432--441, 2008.

\bibitem{malioutov2006walk}
D.~M. Malioutov, J.~K. Johnson, and A.~S. Willsky, ``Walk-sums and belief
  propagation in {Gaussian} graphical models,'' \emph{The Journal of Machine
  Learning Research}, vol.~7, pp. 2031--2064, 2006.

\bibitem{heinemann2012cannot}
U.~Heinemann and A.~Globerson, ``What cannot be learned with bethe
  approximations,'' \emph{arXiv preprint arXiv:1202.3731}, 2012.

\bibitem{ravikumar2011high}
P.~Ravikumar, M.~Wainwright, G.~Raskutti, and B.~Yu, ``High-dimensional
  covariance estimation by minimizing $\ell_{1}$-penalized log-determinant
  divergence,'' \emph{Electronic Journal of Statistics}, vol.~5, pp. 935--980,
  2011.

\bibitem{rothman2008sparse}
A.~Rothman, P.~Bickel, E.~Levina, and J.~Zhu, ``Sparse permutation invariant
  covariance estimation,'' \emph{Electronic Journal of Statistics}, vol.~2, pp.
  494--515, 2008.

\bibitem{johnson2011high}
C.~Johnson, A.~Jalali, and P.~Ravikumar, ``High-dimensional sparse inverse
  covariance estimation using greedy methods,'' \emph{arXiv preprint
  arXiv:1112.6411}, 2011.

\bibitem{mizrahi2013efficient}
Y.~D. Mizrahi, M.~Denil, and N.~de~Freitas, ``Linear and parallel learning of
  {Markov} random fields,'' \emph{arXiv preprint arXiv:1308.6342}, 2013.

\bibitem{massam2013distributed}
H.~Massam and N.~Wang, ``Distributed parameter estimation of discrete
  hierarchical models via marginal likelihoods,'' \emph{arXiv preprint
  arXiv:1310.5666}, 2013.

\bibitem{Petersen06thematrix}
K.~B. Petersen, M.~S. Pedersen, J.~Larsen, K.~Strimmer, L.~Christiansen,
  K.~Hansen, L.~He, L.~Thibaut, M.~Barão, S.~Hattinger, V.~Sima, and W.~The,
  ``The matrix cookbook,'' Tech. Rep., 2006.

\bibitem{murphy1999loopy}
K.~P. Murphy, Y.~Weiss, and M.~I. Jordan, ``Loopy belief propagation for
  approximate inference: An empirical study,'' in \emph{Proceedings of the
  Fifteenth conference on Uncertainty in artificial intelligence}.\hskip 1em
  plus 0.5em minus 0.4em\relax Morgan Kaufmann Publishers Inc., 1999, pp.
  467--475.

\bibitem{pacheco2012minimization}
J.~Pacheco and E.~B. Sudderth, ``Minimization of continuous {Bethe}
  approximations: A positive variation,'' in \emph{Advances in Neural
  Information Processing Systems}, 2012, pp. 2573--2581.

\bibitem{wainwright2006estimating}
M.~J. Wainwright, ``Estimating the wrong graphical model: Benefits in the
  computation-limited setting,'' \emph{The Journal of Machine Learning
  Research}, vol.~7, pp. 1829--1859, 2006.

\bibitem{liang2008asymptotic}
P.~Liang and M.~I. Jordan, ``An asymptotic analysis of generative,
  discriminative, and pseudolikelihood estimators,'' in \emph{Proceedings of
  the 25th international conference on Machine learning}.\hskip 1em plus 0.5em
  minus 0.4em\relax ACM, 2008, pp. 584--591.

\bibitem{van2000asymptotic}
A.~Van~der Vaart, \emph{Asymptotic statistics}.\hskip 1em plus 0.5em minus
  0.4em\relax Cambridge university press, 2000, vol.~3.

\bibitem{wainwright2009sharp}
M.~Wainwright, ``Sharp thresholds for noisy and high-dimensional recovery of
  sparsity using $\ell$1-constrained quadratic programming ({Lasso}),''
  \emph{IEEE Transactions on Information Theory}, vol.~55, no.~5, pp.
  2183--2202, 2009.

\bibitem{meinshausen2006high}
N.~Meinshausen and P.~B{\"u}hlmann, ``High-dimensional graphs and variable
  selection with the lasso,'' \emph{The Annals of Statistics}, vol.~34, no.~3,
  pp. 1436--1462, 2006.

\bibitem{friedman2009elements}
J.~Friedman, T.~Hastie, and R.~Tibshirani, \emph{The Elements of Statistical
  Learning: Data Mining, Inference, and Prediction}.\hskip 1em plus 0.5em minus
  0.4em\relax Springer-Verlag New York, 2009.

\bibitem{guestrin2004distributed}
C.~Guestrin, P.~Bodik, R.~Thibaux, M.~Paskin, and S.~Madden, ``Distributed
  regression: an efficient framework for modeling sensor network data,'' in
  \emph{Information Processing in Sensor Networks, 2004. IPSN 2004. Third
  International Symposium on}.\hskip 1em plus 0.5em minus 0.4em\relax IEEE,
  2004, pp. 1--10.

\bibitem{watts1998small}
D.~Watts and S.~Strogatz, ``The small world problem,'' \emph{Collective
  Dynamics of Small-World Networks}, vol. 393, pp. 440--442, 1998.

\bibitem{wang2010solving}
C.~Wang, D.~Sun, and K.-C. Toh, ``Solving log-determinant optimization problems
  by a newton-cg primal proximal point algorithm,'' \emph{SIAM Journal on
  Optimization}, vol.~20, no.~6, pp. 2994--3013, 2010.

\bibitem{johnson2006fisher}
J.~Johnson, ``Fisher information in {Gaussian} graphical models,''
  \emph{unpublished technical note}, 2006.

\end{thebibliography}
\appendices

\section{Proof of Theorem~\ref{thm:asym}}
\begin{IEEEproof}
\label{app:proofthm}
%(abbreviated) 
Consider the following set of sparse positive semidefinite matrices with respect to a non-zero element set $R$: 
\ALGN{
\mcal{K}^{R} := \{ \mbf{K} \ | \ \mbf{K} \succeq 0, \mbf{K}_{(j,k)} = 0, \forall (j,k) \notin R\}.
}
We first note that, when $R$ is taken to be the relaxed edge set of a neighborhood as defined in~\eqref{eq:setS}, then the true marginal concentration matrix corresponding to the neighborhood, $\mbf{K}^{*} = (\Sigstar_{\mcal{N},\mcal{N}})^{-1}$, must belong to the set $\mcal{K}^{R}$. This can be seen from the fact that the true global concentration matrix $\mbf{J}^{*}$ conforms to the sparsity pattern specified by $\Etilde$ and from relations~\eqref{eq:KequalsJ} and \eqref{eq:KequalsSchur}. Therefore the proposed relaxed MML problem~\eqref{eq:localrelax} is equivalent to a standard ML problem with respect to a GGM distribution parameterized by matrix $\mbf{K} \in \mcal{K}^{R}$, with $\mbf{K}^{*}$ being the population parameter. Then the asymptotic consistency, normality and efficiency of the proposed relaxed MML estimator (with respect to the local problem) all follow from the standard asymptotic analysis of the ML estimator~\cite{van2000asymptotic}. In particular, the variances of the errors achieve  
the diagonal elements of the inverse Fisher information matrix $\mbf{F}$ defined in Eq.~\eqref{eq:fisher} (see~\cite{johnson2006fisher} for the derivation). 
Finally by extracting and summing the variances corresponding to the row parameters, we obtain the expression for the asymptotic mean squared Frobenius error of the proposed global estimator $\widehat{\mbf{J}}^{\text{Relax}}$. 
\end{IEEEproof}

\section{Proof of Theorem~\ref{thm:variance}}
\label{sec:proof:var}

\begin{IEEEproof} 
We first consider the case of $k = 1$, i.e., we compare the asymptotic variances of the one-hop and two-hop RMML estimators.  Subsequently we generalize the arguments to $k > 1$ and to the global ML estimator. Suppressing the index $i$ for local neighborhoods, 
let $\mcal{B}^{j}, \mcal{N}^{j}$ be the sets of buffer and all nodes (i.e. variables) with respect to the $j$-hop neighborhood, respectively ($j = 1,2$).

Next we define some set notation for edge parameters. Let $E^{j}  = \Etilde \cap (\mcal{N}^j \times \mcal{N}^j)$ denote the subset of edges in $\Etilde$ with both endpoints in $\mcal{N}^j$. 
Let $B^{j}$ be the set of all possible edges connecting $j$-hop buffer nodes, i.e. $B^{j} := \mcal{B}^{j} \times \mcal{B}^{j}$. 
Recall  from~\eqref{eq:setL} that $L$ denotes the set of row parameters, which is defined as $L = E^{1} \bs B^{1}$. 
 Finally note that the ($j$-hop) relaxed edge sets defined in Eq.~\eqref{eq:setS} are related to the above two sets as $R^{j} := E^{j} \cup B^{j}$, $j  = 1,2$.

We augment the two-hop neighborhood graph by adding all 
 edges among one-hop buffer nodes and among two-hop buffer nodes that are not already in $E^2$ (see Figure~\ref{fig:chordal} for an illustration). 
This augmented edge set is denoted as $\overline{E^{2}} := E^{2} \cup B^{1} \cup B^{2}$. After this augmentation, the one-hop buffer clique $B^{1}$ separates the two-hop neighborhood graph into two components  and a non-overlapping decomposition follows:
\ALGNN{
\overline{E^{2}} = [\rlap{$\underbrace{L, B^{1},}_{C_{1} = R^{1}}$}L,\overbrace{B^{1}, (E^{2} \bs E^{1}) \cup B^{2}}^{C_{2}}],
}
where we define two subsets $C_{1}$ and $C_{2}$. The augmented two-hop neighborhood graph is therefore decomposed by $(C_{1} \bs B^{1}, B^{1}, C_{2} \bs B^{1})$~\cite[Def. 2.1]{lauritzen1996graphical}.

\begin{figure}[ht]
\centering
  \includegraphics[scale=0.4]{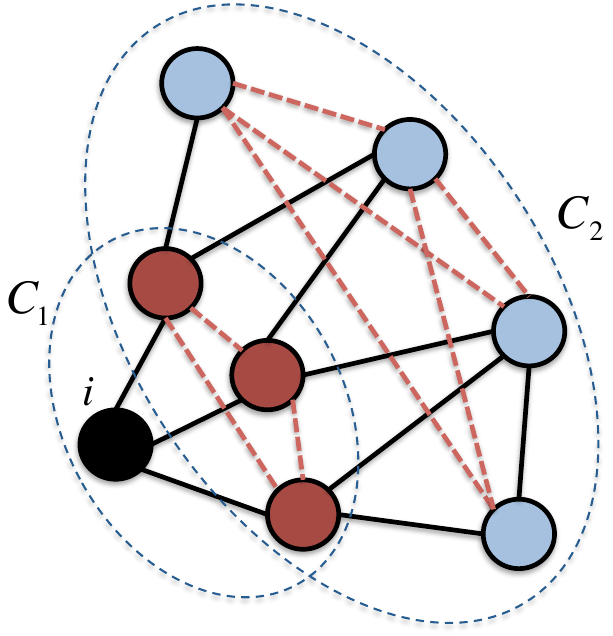}
\caption{Illustration of the graph augmentation in the proof of Theorem~\ref{thm:variance}. Dashed red lines indicate the added edges, and dashed blue contours indicate the sets $C_1$ and $C_2$, which intersect at the one-hop separator clique formed by red nodes.}
\label{fig:chordal}
\end{figure}

Similar to Theorem~\ref{thm:asym}, the asymptotic   error covariance matrix of the RMML estimator for the augmented two-hop neighborhood is the inverse of corresponding Fisher information matrix (FIM), denoted as $\overline{\mbf{F}}$. By Proposition 5.8 in~\cite{lauritzen1996graphical}, the decomposability of the augmented graph leads to the following decomposition of the inverse of FIM:
\ALGN{
\overline{\mbf{F}}^{-1} = \left[ (\overline{\mbf{F}}_{C_{1}, C_{1}})^{-1} \right]^{0} + \left[ (\overline{\mbf{F}}_{C_{2}, C_{2}})^{-1} \right]^{0} - \left[ (\overline{\mbf{F}}_{B^{1}, B^{1}})^{-1} \right]^{0},
}
where $[ \cdot ]^{0}$ appropriately zero-pads its argument to conform to the dimensions of $\overline{\mbf{F}}^{-1}$.

Restricting this relation to the row parameters $L$, we have
\ALGNN{
\overline{\mbf{F}}^{-1}_{L, L} = (\overline{\mbf{F}}_{C_{1}, C_{1}})^{-1}_{L, L},
}
since the row parameters are only contained in $C_{1}$. Noting that set $C_{1}$ is equivalent to the one-hop relaxed edge set $R^{1}$, then
\ALGNN{
\overline{\mbf{F}}^{-1}_{L, L} = (\overline{\mbf{F}}_{C_{1}, C_{1}})^{-1}_{L, L} = (\overline{\mbf{F}}_{R^{1}, R^{1}})^{-1}_{L, L}.
\label{eq:aug-equals-onehop}
}
Therefore, from Theorem~\ref{thm:asym} we have that the asymptotic mean squared error of the RMML estimator using the augmented graph 
is the same as that of the one-hop RMML estimator.

On the other hand, the augmented edge set $\overline{E^{2}}$ is different from the relaxed edge set $R^{2}$ only in the one-hop buffer clique $B^1$. Therefore another possible decomposition of the augmented edge set is (after re-ordering):
\ALGNN{
\overline{E^{2}} = [\underbrace{L, E^{1} \bs L, (E^{2} \bs E^{1}) \cup B^{2}}_{R^{2}}, \ \underbrace{B^{1} \bs E^{1}}_{D}],
}
where we define the difference set as $D$. 
Then using a property of Schur complements of positive semidefinite matrices, the variance matrix corresponding to $R^2$ (i.e. the non-zero pattern of the two-hop RMML estimator) satisfies
\ALGNN{
\overline{\mbf{F}}^{-1}_{R^{2}, R^{2}}  & =  \left( \overline{\mbf{F}}_{R^{2}, R^{2}} - \overline{\mbf{F}}_{R^{2}, D} (\overline{\mbf{F}}_{D, D})^{-1} \overline{\mbf{F}}_{D, R^{2}} \right)^{-1} \notag \\
& \succeq \left( \overline{\mbf{F}}_{R^{2}, R^{2}} \right)^{-1}.	
}
Restricting this relation to the submatrix indexed by set $L$, we have
\ALGNN{
\overline{\mbf{F}}^{-1}_{L,L} \succeq \left( \overline{\mbf{F}}_{R^{2}, R^{2}} \right)^{-1}_{L,L}.	\label{eq:aug-larger-twohop}
}
Now combining Eq.~\eqref{eq:aug-equals-onehop}, Eq.~\eqref{eq:aug-larger-twohop} and Theorem~\ref{thm:asym}, we can conclude that the asymptotic variance of the one-hop RMML estimator (i.e. the mean squared error) is larger than that of the two-hop estimator. 

Similar arguments can be established for comparing the asymptotic variances of the two-hop RMML and the GML estimators, which shows that the asymptotic variance of RMML estimator is larger than that of the GML estimator. The above proof can be easily generalized to arbitrary $k$-hop neighborhoods. 

\end{IEEEproof}

\section{Proof of Theorem~\ref{thm:nonasym}}
\label{sec:proof:nonasym}

The key ingredient in proving Theorem~\ref{thm:nonasym} is the following lemma, which provides a bound for the error of the RMML estimator $\KRelax$~\eqref{eq:localrelax} in a given local neighborhood (the neighborhood index $i$ is suppressed). Let $\Sigstar$ be the true global covariance matrix, and $\Kstar$ be the true marginal precision matrix corresponding to the given neighborhood.

\begin{lemma} \label{prop:localbound}
For a given local neighborhood $\mcal{N}$, if
\ALGNN{
\left\| \left( \Sighat^{(T)} - \Sigstar \right)_{R} \right\|_{\infty} \le \epsilon_{\Sigma} \le \frac{1}{9 \overline{\kappa} \sqrt{|R|}},
\label{eq:lem:sampledeviation}
}
we have
\ALGNN{
\left\| \KRelax - \Kstar \right\|_F \le 9 \overline{\kappa}^2  \epsilon_{\Sigma} \sqrt{|R |}.
}
\end{lemma}

The proof of Lemma~\ref{prop:localbound} is given in Appendix~\ref{sec:proof:prop:local}.
The above lemma is deterministic in nature. To ensure that assumption~\eqref{eq:lem:sampledeviation} is satisfied with high probability when the sample covariance $\Sighat^{(T)}$ is random, 
we make use of the following concentration result for Gaussian random variables by Ravikumar et al.~\cite{ravikumar2011high}:
 \begin{lemma}
For a $p$-dimensional Gaussian random vector with covariance matrix $\Sigstar$, the sample covariance matrix obtained from $T$ samples $\Sighat^{(T)}$ satisfies
\ALGNN{
P\left\{ | \Sighat^{(T)}_{i,j} - \Sigstar_{i,j} | > \epsilon \right\} \le 4 \exp \left( - \frac{T \cdot \epsilon^{2}}{3200 \overline{\sigma}^{2}} \right),
}
for all $\epsilon \in (0, 40 \overline{\sigma})$, where 
%\ALGNN{
$\overline{\sigma} := \max_{i=1,\ldots,p} \Sigstar_{i,i}$.
%}
\label{lem:sampledeviation}
\end{lemma}

Now we are ready to prove Theorem~\ref{thm:nonasym}.
\begin{IEEEproof} (Theorem~\ref{thm:nonasym})
Given the condition~\eqref{eq:samplecomplexity1} on $T$, we have 
\ALGNN{
C \sqrt{\frac{3200 \overline{\sigma}^{2} \log p^2 }{T}} \le 40 \overline{\sigma}.
}
Then applying Lemma ~\ref{lem:sampledeviation} and the union bound, we have
\ALGNN{
\begin{split}
P & \left\{ \left\| \left( \Sighat^{(T)} - \Sigstar \right)_{R_{i}} \right\|_{\infty} \le C \sqrt{\frac{3200 \overline{\sigma}^{2} \log p^2}{T}} \right\} \\
& \ge P \left\{ \left\| \Sighat^{(T)} - \Sigstar  \right\|_{\infty} \le C \sqrt{\frac{3200 \overline{\sigma}^{2} \log p^2}{T}} \right\} \\
& \ge 1 - \frac{4}{p^{2(C^2 - 1)}}.
\end{split}
\label{eq:event}
}
Conditioned on the event in~\eqref{eq:event},  condition~\eqref{eq:samplecomplexity1} also guarantees that~\eqref{eq:lem:sampledeviation}   holds for all local neighborhoods.  Then the total Frobenius error in the global estimate $\JRelax$ can be bounded  by Lemma~\ref{prop:localbound}:
\ALGN{
\| \JRelax - \Jstar \|_F  & \overset{(i)}{=} 
\left( \sum_{i = 1}^{p} \| (\JRelax - \Jstar)_{L_{i}} \|^{2}_F \right)^{1/2} \\
& \overset{(ii)}{=} \left( \sum_{i = 1}^{p} \| (\KRelax - \Kstar)_{L_{i}} \|^{2}_F \right)^{1/2} \\
& \overset{(\text{Lem.~\ref{prop:localbound}})}{\le} \left( \sum_{i = 1}^{p} \left( 9 \overline{\kappa}^2  C \sqrt{\frac{3200 \overline{\sigma}^{2} |R_{i} | \log p^2}{T}} \right)^2  \right)^{1/2} \\
& \le 720 C \cdot \overline{\kappa}^2  \overline{\sigma}  \sqrt{\frac{r \log p }{T}},
}
where  identity $(i)$ is due to the fact that the global estimator is a concatenation of non-overlapping row parameter sets (see Eq.~\eqref{eq:setL} for definition of $L_i$'s), equality $(ii)$ is due to our construction of $\JRelax$ from $\KRelax$ (see Eq.~\eqref{eq:estimatorRMMLE}), and the fact that row parameters are always protected. 
\end{IEEEproof}

\section{Proof of Lemma~\ref{prop:localbound}}
\label{sec:proof:prop:local}
\begin{IEEEproof}
The main idea of this proof is inspired by~\cite{rothman2008sparse}. The difference is that we focus on the local RMML problem, rather than the global ML problem (which is studied in~\cite{rothman2008sparse}). Define the marginal likelihood function for a local neighborhood $\mcal{N}$ as $\mcal{L}(\mbf{K}) = \langle \Sighat_{\mcal{N}, \mcal{N}}^{(T)}, \mbf{K} \rangle - \log \det (\mbf{K})$, where we super-script the sample covariance to emphasize that it is obtained from $T$ samples.  

Recall $\Kstar := \left( \Sigstar_{\mcal{N}, \mcal{N}} \right)^{-1}$ is the local marginal precision matrix. Define the shorthand notation for the local RMML estimate as $\Khat := \KRelax$.

Consider the function $\mcal{Q}(\Del) := \mcal{L}(\Kstar + \Del) - \mcal{L}(\Kstar)$, where $\Del$ respects the sparsity structure of the RMML problem, i.e. $\Del_{R^C} = \mbf{0}$ and $\Del = \Del^T$. 
Let $0 < \delta \le \overline{\kappa}$ be a given radius, define the following set
\ALGNN{
\mcal{C}(\delta) := \{ \Del \ | \ \Del_{R^C} = \mbf{0}, \Del = \Del^T, \| \Del \|_{F} = \delta \}, % \sqrt{\frac{|R| \log |R|}{T}} \},
\label{eq:setC}
}
where $R$ is the local relaxed edge set. Note that $\mcal{C}(\delta)$ defines a sphere, not a ball.

Note that $\mcal{Q}(\Del)$ is a convex function of $\Del$. By construction we have $\mcal{Q}(\mbf{0}) = 0$, and the optimality of $\KRelax$ implies that $\mcal{Q}(\Deltahat) \le \mcal{Q}(\mbf{0}) = 0$, where we define $\Deltahat := \Khat - \Kstar$. 
Then if we can establish that
\[
\inf_{\Del \in \mcal{C}(\delta)} \mcal{Q}(\Del) > 0,
\]
then the optimal error matrix $\Deltahat$ must lie inside the sphere defined by $\mcal{C}(\delta)$ by convexity of $\mcal{Q}$, implying that 
$\| \Deltahat \|_F \le \delta$.
Now it suffices to find a suitable radius $\delta > 0$ such that $\mcal{Q}(\Del)$ is lower-bounded from zero for all $\Del \in \mcal{C}(\delta)$.

Since
\ALGNN{
\mcal{Q}(\Del) & = \mcal{L}(\Kstar + \Del) - \mcal{L}(\Kstar) \notag \\
& = \langle \Sighat_{\mcal{N}, \mcal{N}}^{(T)}, \Del \rangle - ( \log \det (\Kstar + \Del) - \log \det (\Kstar)). \notag
}
Similar to~\cite{rothman2008sparse}, we make use of the Taylor's theorem for the $\log \det(\cdot)$ function
\ALGNN{
& \log \det (\Kstar + \Del) - \log \det (\Kstar)  = \langle (\Kstar)^{-1}, \Del \rangle - \notag \\
&\overrightarrow{\Del}^T \left[ \int_0^1 (1-t) (\Kstar + t \Del)^{-1} \otimes (\Kstar + t \Del)^{-1} dt \right] \overrightarrow{\Del},
}
where $\otimes$ denotes the Kronecker product, and $\overrightarrow{\Del}$ is the properly vectorized form of matrix $\Del$.

Using this identity, we have
\ALGNN{
\mcal{Q}(\Del) & = \underbrace{\langle \Sighat_{\mcal{N}, \mcal{N}}^{(T)} - (\Kstar)^{-1}, \Del \rangle}_{T_1} + \notag \\
& \underbrace{\overrightarrow{\Del}^T \left[ \int_0^1 (1-t) (\Kstar + t \Del)^{-1} \otimes (\Kstar + t \Del)^{-1} dt \right] \overrightarrow{\Del}}_{T_2}.
\label{eq:T1T2}
}
Next we bound $T_1$ and $T_2$ defined above separately.

For $T_1$, notice that the difference matrix $\Del$ is non-zero only in a restricted set $R$, therefore it reduces to a lower-dimensional inner product:

\ALGNN{
\begin{split}
| T_1 | & = | \langle ( \Sighat^{(T)} - \Sigstar )_R, \Del_R \rangle | \\
& \overset{(i)}{\le} \| ( \Sighat^{(T)} - \Sigstar )_R \|_\infty \cdot \| \Del_R \|_1 \\
& \overset{\text{Eq.\eqref{eq:lem:sampledeviation}}}{\le} \epsilon_{\Sigma} \cdot \sqrt{| R |} \cdot \| \Del \|_F,
\end{split}
\label{eq:boundT1}
}
where $(i)$ is due to the duality between norms $\| \cdot \|_\infty$ and $\| \cdot \|_1$.

For $T_2$, we follow similar derivations as in~\cite{rothman2008sparse}:
\ALGNN{
\begin{split}
T_2 & \ge \| \Del \|_F^2 \cdot \la_{\min} \left( \int_0^1 (1-t) (\Kstar + t \Del)^{-1} \otimes (\Kstar + t \Del)^{-1} dt \right) \\
& \overset{(i)}{\ge} \| \Del \|_F^2 \int_0^1 (1-t) \la_{\min}^2 \left( (\Kstar + t \Del)^{-1} \right) dt \\
& \ge \frac{1}{2} \| \Del \|_F^2 \min_{0 \le t \le 1} \la_{\min}^2 \left( (\Kstar + t \Del)^{-1} \right) \\
& \overset{(ii)}{\ge} \frac{1}{2} \| \Del \|_F^2 \min_{\| \widetilde{\Del} \|_F \le \delta  } \la_{\min}^2 \left( (\Kstar + \widetilde{\Del})^{-1} \right) \\
& \ge \frac{1}{2} \| \Del \|_F^2 \min_{\| \widetilde{\Del} \|_F \le \delta } \| \Kstar + \widetilde{\Del} \|_2^{-2} \\
& \ge \frac{1}{2} \| \Del \|_F^2 \min_{\| \widetilde{\Del} \|_F \le \delta } (\| \Kstar \|_2 + \| \widetilde{\Del} \|_2 )^{-2} \\
& \overset{(iii)}{\ge} \frac{1}{2}  \| \Del \|_F^2 \min_{\| \widetilde{\Del} \|_F \le \delta  } ( \overline{\kappa}  + \| \widetilde{\Del} \|_F )^{-2} \\
& = \frac{1}{8 \overline{\kappa}^{2}  } \| \Del \|_F^2,
\end{split}
\label{eq:boundT2}
}
where $(i)$ follows the eigenvalue property of Kronecker product, $(ii)$ is due to the fact that $\Del \in \mcal{C}(\delta)$, $(iii)$ is due to the interlacing property of eigenvalues of sub-matrices 
\ALGNN{
\begin{split}
\| \Kstar \|_{2} & = 
\frac{1}{\la_{\min} (\Sigstar_{\mcal{N}, \mcal{N}})} \le \frac{1}{\la_{\min} (\Sigstar)} = \| \Jstar \|_{2} = \overline{\kappa},
\end{split}
\label{eq:spectralK}
}
The last inequality is due to construction, i.e. $\delta \le \overline{\kappa}$.

Now $\mcal{Q}(\Del)$ can be bounded by% (with high probability)
\ALGNN{
\mcal{Q}(\Del) & \ge -  \epsilon_{\Sigma} \cdot \sqrt{| R |} \cdot \| \Del \|_F + \frac{1}{8 \overline{\kappa}^{2}  } \| \Del \|_F^2 \\ 
& =  \| \Del \|_F \left( \frac{1}{8 \overline{\kappa}^{2}  } \| \Del \|_F - \epsilon_{\Sigma} \cdot \sqrt{| R |} \right).
}

The proof is complete if the RHS can be lower bounded away from zero.
It can be verified that with the choice of $\epsilon_{\Sigma}$ as in~\eqref{eq:lem:sampledeviation}, letting $\delta = 9 \overline{\kappa}^{2} \epsilon_{\Sigma} \sqrt{| R |}$ suffices. Therefore $\| \Deltahat \|_{F} \le \delta = 9 \overline{\kappa}^{2} \epsilon_{\Sigma} \sqrt{| R |}$.

\end{IEEEproof}

\end{document}